%% file: root.tex
\documentclass[journal,letterpaper]{IEEEtran}

\IEEEoverridecommandlockouts                              

\usepackage{amsthm}
\usepackage{times}
\usepackage{multicol}
\usepackage[bookmarks=true]{hyperref}
\usepackage{xcolor}
\usepackage{hyperref}
\usepackage{amsmath, amssymb}
\usepackage{amsfonts}
\usepackage{graphicx}
\usepackage{siunitx}
\usepackage{standalone}
\usepackage{booktabs}
\usepackage[ruled,vlined,linesnumbered,noend]{algorithm2e}
\usepackage{mdframed}
\usepackage{fancyvrb,multirow}
\usepackage{soul}
\usepackage{dsfont,mathabx}
\usepackage{array, booktabs}
\usepackage{subfigure}
\usepackage{booktabs}
\usepackage{makecell}
\usepackage{tabularx}
\usepackage{booktabs}
\usepackage{amsmath,amsfonts,amsthm,bm} %
\usepackage[para,online,flushleft]{threeparttable}
\usepackage[T1]{fontenc}
\usepackage{tikz}
\usepackage{pgfplotstable}
\usepackage{pgfplots}

\newtheorem{theorem}{Theorem}

\newtheorem{lemma}{Lemma}

\theoremstyle{definition}

\newtheorem{definition}{Definition}
\newtheorem{remark}{Remark}

\newtheorem{problem}{Problem}

\newcommand{\cb}{\color{black}}

\title{\LARGE \bf PushingBots: Collaborative Pushing via
  Neural Accelerated Combinatorial Hybrid Optimization}

\author{Zili Tang, Ying Zhang and Meng Guo
  \thanks{The authors are with the School of Advanced Manufacturing and Robotics,
    Peking University, Beijing 100871, China.
This work was supported by the National Key Research and Development Program of China under grant
2023YFB4706500; the National Natural Science Foundation
of China (NSFC) under grants 62203017, U2241214, T2121002.

Corresponding author: Meng Guo, {\tt\small meng.guo@pku.edu.cn}.}
}

\begin{document}
\maketitle
\thispagestyle{empty}
\pagestyle{empty}

\input{contents/abstract.tex}
\input{contents/introduction.tex}
\input{contents/related.tex}
\input{contents/problem.tex}
\input{contents/solution.tex}
\input{contents/experiment.tex}
\input{contents/conclusion.tex}
\input{contents/ack.tex}
\bibliographystyle{IEEEtran}
\bibliography{contents/references}
\input{contents/appendix.tex}
\end{document}

%% file: contents/abstract.tex
\begin{abstract}
  Many robots are not equipped with a manipulator
  and many objects are not suitable for prehensile manipulation (such as large boxes and cylinders).
  In these cases, pushing is a simple yet effective non-prehensile skill for robots 
  to interact with and further change the environment.
  Existing work often assumes a set of predefined pushing modes and fixed-shape objects.
  This work tackles the general problem
  of controlling a robotic fleet to push collaboratively
  numerous arbitrary objects to respective destinations,
  within complex environments of cluttered and movable obstacles.
  It incorporates several characteristic challenges for
  multi-robot systems such as online task coordination
  under large uncertainties of cost and duration,
  and for contact-rich tasks
  such as hybrid switching among different contact modes,
  and under-actuation due to constrained contact forces.
  The proposed method is based on combinatorial hybrid optimization
  over dynamic task assignments and hybrid execution
  via sequences of pushing modes and associated forces.
  It consists of three main components:
  (I) the decomposition, ordering and rolling assignment
  of pushing subtasks to robot subgroups;
  (II) the keyframe guided hybrid search to
  optimize the sequence of parameterized pushing modes
  for each subtask;
  (III) the hybrid control to execute these modes and transit among them.
  Last but not least, a diffusion-based accelerator
  is adopted to predict the keyframes and pushing modes that should be
  prioritized during hybrid search;
  and further improve planning efficiency.
  The framework is complete under mild assumptions.
  {Its efficiency and effectiveness under different numbers
  of robots and general-shaped objects are validated extensively
  in simulations and hardware experiments},
  as well as generalizations to heterogeneous robots, planar assembly
  and 6D pushing.
\end{abstract}

%% file: contents/introduction.tex
\section{Introduction}\label{sec:intro}

\begin{figure}[t!]
  \centering
  \includegraphics[width=1\linewidth]{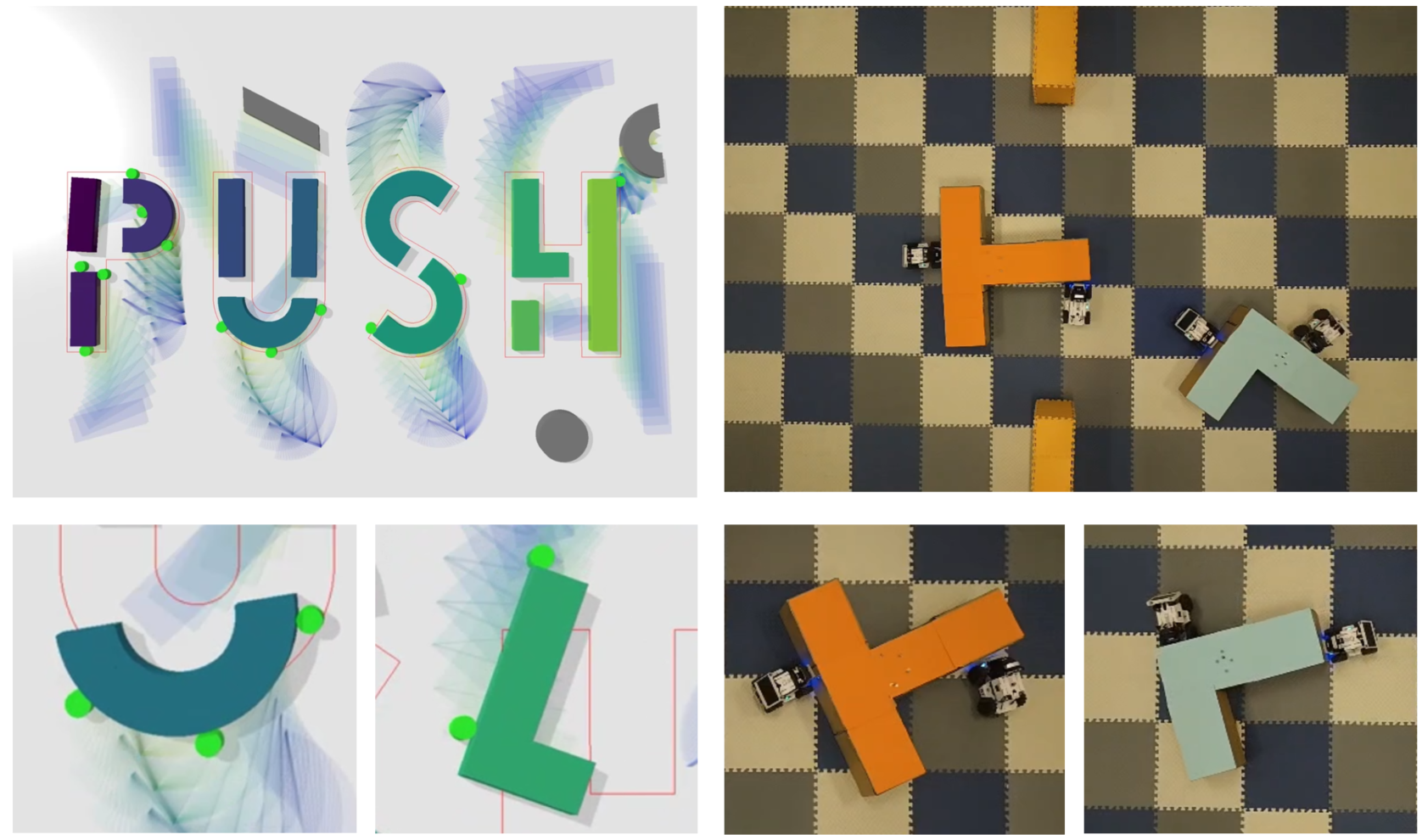}
  \vspace{-5mm}
  \caption{Snapshots of the PushingBots system,
    during the simulated planar assembly task via~$12$
    robots and~$14$ objects (\textbf{Left});
    and the hardware experiments of swapping~$2$ objects via~$4$
    mini-ground vehicles (\textbf{Right}).}
  \label{fig:intro}
  \vspace{-5mm}
\end{figure}

Humans often interact with objects via non-prehensile skills
such as pushing and rolling, especially when prehensile
skills such as stable grasping is infeasible.
This aspect is however less exploited in robotic systems.
{Most existing work treats pushing as a complementary skill
to pick-and-place primitives for a single manipulator
within simple environments,
e.g.,~\cite{goyal1989limit, lynch1992manipulation, hogan2020reactive, xue2023guided}.
Nonetheless, pushing can be particularly beneficial
for low-cost mobile robots that are {not} equipped with a manipulator,
e.g., ground vehicles,
quadruped robots,
and even underwater vehicles~\cite{jeon2023learning}.
For instance, obstacles can be pushed out of the path,
and target objects can be pushed to desired positions.
In addition, several robots could improve the feasibility and efficiency by
collaboratively and simultaneously pushing the same object
at different points with different forces,
which might be otherwise too heavy for individual robots.
On a larger scale, as shown in Fig.~\ref{fig:intro},
a fleet of such robots can be deployed to transport concurrently
numerous objects,
yielding a much higher capacity and reliability.}

Despite its intuitiveness, the collaborative pushing task imposes
several challenges that are typical for contact-rich
tasks~\cite{toussaint2015logic}.
Different motions of the object, such as translation and rotation,
can be generated by different modes of pushing such
as long-side, short-side, diagonal, and caging,
however with varying quality depending on the physical properties
of the object~\cite{goyal1989limit},
e.g., mass distribution, shape, and friction coefficients.
Thus, long-term pushing in complex environments would require:
(I) planning under tight kinematic and geometric constraints
such as narrow passages and sharp turns~\cite{hogan2020reactive};
(II) hybrid optimization to switch
between different contact modes with desired pushing
forces~\cite{toussaint2018differentiable};
and (III) underactuated control due to constrained contact forces
and unmodeled friction or slipping~\cite{zeng2018learning,xiao2020learning},
yielding a difficult problem not only for
modeling and planning, but also for hybrid control.
Lastly, the coordination of multiple robots to push numerous objects
is also non-trivial, due to:
(I) the tight coupling of object trajectories,
both spatial and temporal to avoid collisions and
deadlocks~\cite{garcia2013scalable, kube1997task};
and (II) the inherent uncertainties in the
feasibility and duration of subgroups pushing different
objects during online execution~\cite{vig2006multi}.

\begin{figure*}[t!]
  \centering
  \includegraphics[width=0.95\linewidth]{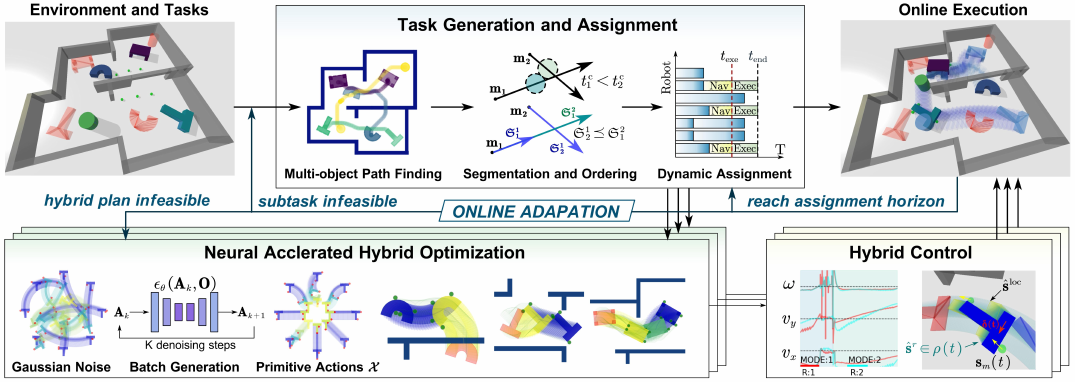}
  \caption{{Overall proposed online planning
      and adaptation framework,
      consisting of the MAPF-based
      task decomposition, subtask generation
      with partial ordering,
    the online rolling assignment of subtasks,
    the neural accelerated hybrid optimization,
    and the online hybrid control.}}
  \label{fig:overall}
  \vspace{-5mm}
\end{figure*}

To tackle these challenges,
this work addresses the problem of controlling mobile robots to
collaboratively push arbitrary-shaped objects to goal positions
in a complex environment without predefined contact modes.
As illustrated in Fig.~\ref{fig:overall},
the overall problem is formulated as a combinatorial hybrid optimization (CHO)
including the dynamic assignment of robots to different objects,
and the hybrid optimization of pushing modes and forces for each object.
To begin with,
the pushing task of all objects is decomposed into a set of temporally-ordered
subtasks via the multi-agent path finding (MAPF) algorithm for
irregular-shaped vehicles and spatial segmentation.
Then, to handle the uncertain execution time of each subtask,
a receding-horizon coordination algorithm is proposed to dynamically
assign the robots to different subtasks based on the estimated feasibility and duration.
Moreover,
a learning-while-planning hybrid search algorithm
iteratively decomposes subtasks via keyframe sampling
and generates suitable modes to minimize feasibility loss,
transition cost, and control efforts.
Lastly, the online execution and adaptation strategy is presented
for both the task assignment and the hybrid control.
The diffusion-based accelerator is learned offline and online,
to predict the sequence of keyframes
and pushing modes for different subtasks,
along with the estimated cost.
Theoretical guarantees on completeness and performance are ensured
under mild assumptions.
Extensive simulations and hardware experiments are conducted
to validate its performance and
the effectiveness of the learned accelerator.

The main contributions of this work are two-fold:
(I) the novel combinatorial-hybrid optimization algorithm
with neural acceleration,
for the multi-robot planar pushing problem of arbitrary objects,
without any predefined contact modes or primitives;
(II) the theoretical condition for feasibility and completeness,
w.r.t. an arbitrary number of robots and general-shaped objects.
To the best of our knowledge, this is the first work that
provides such results.

%% file: contents/related.tex
\section{Related Work}\label{sec:related}
\subsection{Task and Motion Planning}\label{subsec:tmp}
The predominant direction in task and motion planning focuses on navigation and prehensile manipulation, particularly on grasping strategies and object manipulation sequences~\cite{pan2022algorithms}.
Representative tasks include sequential assembly~\cite{toussaint2015logic, guo2021geometric}, rearrangement~\cite{kim2019learning}, and structural construction~\cite{22-hartmann-TRO}.
In contrast, non-prehensile planar manipulation poses unique challenges due to the lack of stable grasps.
The classical single-pusher-single-slider model~\cite{goyal1989limit} highlights the importance of online adaptation of discrete contact decisions and force constraints, often resulting in exponential planning complexity.
To manage this, motion primitives such as sticking and sliding can be encoded into integer programs~\cite{hogan2020reactive}, and various search-based optimization approaches have been proposed~\cite{toussaint2018differentiable, wang2021learning}.
However, most existing methods focus on single manipulators and fixed-shape objects in simple settings, limiting their applicability to multi-robot, multi-object scenarios.
Coordinating multiple robots for multi-object tasks introduces significant complexity.
Typically, tasks must be decomposed into subtasks with well-defined dependencies,
which are subsequently assigned to robot subgroups,
as comprehensively surveyed in~\cite{torreno2017cooperative, gini2017multi, khamis2015multi}.
Since many task planning problems are NP-hard or NP-complete~\cite{torreno2017cooperative},
meta-heuristic methods such as local search and genetic algorithms~\cite{khamis2015multi} are commonly employed to improve computational efficiency.
However, these approaches often assume deterministic costs, limiting their applicability in dynamic environments where task outcomes are uncertain and evolve over time.
Recent work has addressed the multi-robot scheduling problem under ordered and uncertain tasks via the graph-based learning~\cite{wang2022heterogeneous},
distributed coordination~\cite{guo2018multirobot},
and search-based adaptation~\cite{choudhury2022dynamic},
demonstrating strong performance in dynamic settings.
These methods operate on the abstract task graphs without incorporating the underlying physical constraints.
In contrast, the decomposition and ordering of the pushing tasks for different objects in this work are not predefined,
rather evaluated online based on the assigned robot subteams and interaction constraints.

\subsection{Collaborative Pushing}\label{subsec:pushing}
As an application of multi-robot systems, collaborative object transportation~\cite{tuci2018cooperative}
can be realized via pushing, grasping, lifting, pulling, or caging.
Pushing is particularly appealing since it often requires no additional hardware:
robots can interact with objects using any part of their body.
Early heuristic strategies, such as hand-tuned state machines for a few robots in free space~\cite{kube1997task}
or occlusion-based pushing rules for convex objects~\cite{chen2015occlusion},
do not generalize well to cluttered or more complex environments.
A hybrid optimization problem is formulated
in~\cite{tang2023combinatorial} for several objects,
but with a set of predefined pushing primitives and cost estimation.
Similarly, sequential pushing steps are manually defined
for heterogeneous robots in~\cite{vig2006multi} to clear
movable rectangular boxes in the workspace,
where only one robot is assigned for each box.
Recent work~\cite{sombolestan2023hierarchical} adopts several
quadruped robots to push objects of unknown mass and friction,
which however allows a limited set of pushing modes.
Another line of work addresses a similar problem,
but only for a single object in free~\cite{rosenfelder2024force,ebel2024cooperative}
or in cluttered~\cite{tang2024collaborative,EbelEberhard2022} space.
Current pushing controllers broadly fall into two categories:
{force-based} and {kinematic}.
The work in~\cite{rosenfelder2024force} exemplifies the former.
Our method, along with~\cite{chen2015occlusion, ebel2024cooperative, EbelEberhard2022}, belongs to the latter,
relying on a kinematic controller without force sensing.
Separately from the control paradigm, prior work also differs in kinematic constraints.
Both~\cite{chen2015occlusion,ebel2024cooperative} use differential-drive robots,
where~\cite{ebel2024cooperative} explicitly considers nonholonomic constraints by Jourdain's principle.
By contrast, our work, together with~\cite{rosenfelder2024force,EbelEberhard2022},
targets holonomic platforms; nonetheless, our pipeline can be partially generalized to nonholonomic settings.
Furthermore, several works adopt distributed schemes for collaborative pushing~\cite{rosenfelder2024force,ebel2024cooperative,EbelEberhard2022},
which optimize interaction modes online via distributed coordination and control,
enabling accurate path tracking in free space~\cite{rosenfelder2024force,ebel2024cooperative}
and even in arbitrarily cluttered environments~\cite{EbelEberhard2022},
the latter being the most similar to our task setup but considers only a single object.
In contrast, this work adopts a centralized framework to tackle multi-robot,
multi-object pushing in cluttered environments,
with a strong focus on multi-object spatiotemporal coordination, online adaptation and neural acceleration.

\subsection{Neural Policies for Task and Motion Planning}\label{subsec:neural}
Due to its uncertainty and complexity,
pushing a T-block has been one of the standard benchmarks
in neural motion policies,
e.g., self-supervised reinforcement learning~\cite{zeng2018learning},
implicit behavioral cloning~\cite{florence2021implicit}
and diffusion-based methods~\cite{chi2023diffusionpolicy}.
Impressive precision and robustness have been shown using only visual inputs.
Similar neural policies are applied to a quadruped robot
for planar pushing of regular objects with whole-body motion control~\cite{jeon2023learning}.
For long-term tasks that require switching among different objects and tools,
neural policies are proposed to predict the contact sequence~\cite{simeonov2021long}
and the sequence of action primitives in~\cite{kim2019learning,driess2021learning}.
However, these aforementioned works mostly
rely on a \emph{single} manipulator
or quadruped robot via one-point pushing,
and human demonstrations.
For the multi-robot setup in this work,
a simultaneous multi-point pushing strategy
is required, making it difficult for humans
to provide such high-quality demonstrations.
Furthermore,
due to the multi-modality of both contact modes and pushing strategies,
the long-term pushing task in complex environments is particularly
challenging for the traditional methods of imitation learning~\cite{mandlekar2020learning}.
Diffusion models on the other hand have shown great potential in modeling such multi-modality,
i.e., as trajectory samplers for online motion planning~\cite{chi2023diffusionpolicy, carvalhomotion}.
However, the planning efficiency and generalization
of the resulting policy have not been fully explored in
multi-body systems with coupled and switching dynamics.
Lastly, the performance guarantee of most aforementioned methods
relies on the test scenario to be close to the training datasets,
which however may not always hold and can lead to failures during online execution~\cite{mandlekar2020learning}.
Thus, it remains challenging to provide formal verification and performance guarantees when executing such learned policies.

%% file: contents/problem.tex

\section{Problem Description}\label{sec:problem}

\subsection{Model of Workspace and Robots}\label{subsec:ws}

Consider a team of~$N$ robots~$\mathcal{R}\triangleq \{R_n,\,n\in \mathcal{N}\}$
that collaborate in a shared 2D workspace~$\mathcal{W}\subset \mathbb{R}^2$,
where~$\mathcal{N}\triangleq \{1,\cdots,N\}$.
The robots are homogeneous, each with a circular or rectangular footprint.
The state of each robot~$R_n$ at time~$t\geq 0$ is defined by 
$\mathbf{s}_n(t)\triangleq (\mathbf{x}_n(t),\psi_n(t))$,
where $\mathbf{x}_n(t)\!\in\!\mathbb{R}^2$ is its position
and $\psi_n(t)\!\in\!(-\pi,\pi]$ its orientation.
The associated generalized velocity is
$\mathbf{p}_n(t)\triangleq\bigl(\mathbf{v}_n(t),\omega_n(t)\bigr)$
with its linear velocity $\mathbf{v}_n(t)\!\in\!\mathbb{R}^2$
and angular velocity $\omega_n(t)\!\in\!\mathbb{R}$.
Let $R_n(t)\subset\mathcal{W}$ denote the area occupied by robot~$R_n$ at time~$t$.
Moreover, each robot tracks the desired velocity 
$\mathbf{u}_n(t)\triangleq (\widehat{\mathbf{v}}_n(t),\widehat{\omega}_n(t))$ 
with a low-level feedback controller, 
which induces second-order dynamics driven by the controller and external contact forces.
Let~$\mathbf{S}_{\mathcal{N}}(t)\triangleq \{\mathbf{s}_n(t)\}$ denote
the state of all robots.
In addition, the workspace is cluttered with a set of obstacles, 
denoted by~$\mathcal{O} \subset \mathcal{W}$.
Thus, the free space is defined
by~$\widehat{\mathcal{W}}\triangleq \mathcal{W}\backslash \mathcal{O}$.

Lastly, there is a set of~$M>0$ target objects~$\boldsymbol{\Omega}
\triangleq \{\Omega_1,\cdots,\Omega_M\} \subset \widehat{\mathcal{W}}$,
where the shape of each object~$\Omega_m$ is an arbitrary polygon formed by
$V_m\geq 3$ ordered vertices ${p_1p_2\cdots p_{V_m}}$,~$\forall
m\in \mathcal{M}\triangleq \{1,\cdots,M\}$.
{Similarly,
the state of each object~$\Omega_m$ at time~$t\geq 0$ is defined by
$\mathbf{s}_m(t)\triangleq (\mathbf{x}_m(t),\psi_m(t))$,
and the associated generalized velocity is
$\mathbf{p}_m(t)\triangleq\bigl(\mathbf{v}_m(t),\omega_m(t)\bigr)$,
with its position~$\mathbf{x}_m(t)\!\in\!\mathbb{R}^2$,
orientation~$\psi_m(t)\!\in\!(-\pi,\pi]$,
linear velocity~$\mathbf{v}_m(t)\!\in\!\mathbb{R}^2$,
and angular velocity~$\omega_m(t)\!\in\!\mathbb{R}$.
Its occupied area at time $t$ is denoted by~$\Omega_m(t)\subset\widehat{\mathcal{W}}$.
}
In addition, its physical parameters are known a priori, 
including its mass~$\mathsf{M}_m$,
the moment of inertia~$\mathsf{I}_m$,
the pressure distribution at the bottom surface,
the coefficient of lateral friction~$\mu^{\texttt{c}}_{m}>0$,
and the coefficient of ground friction~$\mu^{\texttt{s}}_{m}>0$.

\begin{remark}\label{remark:static}
  The obstacles are assumed to be static and known above
  for simplicity,
  which can be relaxed to be movable, unknown or dynamic,
  as discussed in the sequel.
  \hfill $\blacksquare$
\end{remark}

\subsection{{Collaborative Pushing Modes}}
\label{ss:interaction_mode}
\def\sss{\scriptscriptstyle{}}
{The robots can collaboratively push the objects by
making contacts with the objects at different boundary contact points,
called \emph{collaborative pushing modes}.}
More specifically,
as is illustrated in Fig.~\ref{fig:intro} and~\ref{fig:overall},
{a pushing mode for object~$\Omega_m\in \mathcal{M}$ is defined
by~$\boldsymbol{\xi}_m\triangleq (\xi_m,\, \mathbf{F}_m,\,\mathcal{N}_m)$,
where (I) $\mathcal{N}_m\triangleq \{1,\cdots,N_m\}
\subseteq \mathcal{N}$ is the subgroup of robots assigned to push object~$\Omega_m$;
(II) $\xi_m \triangleq \mathbf{c}_1\mathbf{c}_2\cdots\mathbf{c}_{N_m}$
with~$\mathbf{c}_n\in \partial \Omega_m$ being the contact point on the boundary
of the object for the $n$-th robot,
$\forall n\in \mathcal{N}_m$;
and (III) $\mathbf{F}_m \triangleq \mathbf{f}_1\mathbf{f}_2\cdots\mathbf{f}_{N_m}$
with~$\mathbf{f}_n\in \mathbb{R}^2$ being the contact force
exerted by the $n$-th robot.}
Since the set of contact points are \emph{not} pre-defined,
the complete set of all pushing modes
is potentially infinite, denoted by~$\Xi_m$.
Thus, under different pushing modes,
robots can apply different pushing forces
at different contact points, 
leading to distinct object motions (translation and rotation).

\begin{remark}\label{remark:dynamics}
  {Detailed modeling of the coupled dynamics during
    pushing,  including the decomposition of
    the interaction forces and frictions,
    can be found in the Appendix~\ref{app:model}.}
  \hfill $\blacksquare$
\end{remark}

\subsection{Problem Statement}\label{subsec:objective}
The planning objective is to compute a {hybrid plan},
including an object trajectory~$\mathbf{s}_m(t)$,
a sequence of pushing modes~$\mathbf{\xi}_m(t)$,
the pushing robots~$\mathcal{N}_m(t)$,
the required control~$\mathbf{u}_{N_m}$,
such that each object~$\Omega_m\in \mathcal{M}$ is moved from a given
initial state~$\mathbf{s}^{\texttt{0}}_m$
to a given goal state~$\mathbf{s}_m^{\texttt{G}}$, $\forall m \in \mathcal{M}$.
Meanwhile, the robots and the objects should avoid collision
with each other and with all obstacles at all times.
More precisely, it is formulated as a combinatorial hybrid optimization (CHO) problem
as follows:
$$
\underset{\substack{T, \{\mathbf{u}_{n}(t),\forall n\},\\
      \{(\mathbf{s}_m(t),\boldsymbol{\xi}_m(t)),\forall m\}}}
  {\textbf{min}}
  \Big{\{}T+
  \alpha \sum_{t\in \mathcal{T}}\sum_{m\in \mathcal{M}}
      \mathrm{J}_m\big(\boldsymbol{\xi}_m(t),\mathbf{S}_{\mathcal{N}}(t),\mathbf{s}_m(t)\big)
  \Big{\}}
  $$
\vspace{-0.1in}
\begin{equation}\label{eq:problem}
\begin{split}
&\textbf{s.t.}\quad \mathbf{s}_m(0)= \mathbf{s}_m^{\texttt{0}},\;
\mathbf{s}_m(T)= \mathbf{s}_m^{\texttt{G}},\;\forall m; \\
& \quad \quad \mathcal{N}_{m_1}(t) \cap \mathcal{N}_{m_2}(t) = \emptyset,\\
& \quad \quad \Omega_{m_1}(t) \cap \Omega_{m_2}(t) = \emptyset,
\;\forall m_1\neq m_2,\; \forall t;\\
& \quad \quad \, \Omega_m(t)\subset\widehat{\mathcal{W}},\,
  R_n(t)\subset\widehat{\mathcal{W}},\; \forall m,\, \forall n,\; \forall t; \\
\end{split}
\end{equation}
where~$T>0$ is the task duration to be optimized as the largest over all objects;
$\mathcal{T}\triangleq \{0,1,\cdots,T\}$;
$\forall t$, $\forall m$ and $\forall n$ are short for $\forall t\in \mathcal{T}$,
$\forall m\in \mathcal{M}$, $\forall n\in \mathcal{N}$, respectively;
the cost function~$\mathrm{J}_m:\Xi_m \times \mathbb{R}^{3N} \times \mathbb{R}^{3} \rightarrow \mathbb{R}_{+}$ is
a given function to measure the feasibility, stability, robustness and control cost,
of choosing a certain mode and the forces given the desired object trajectory,
which is different among the objects due to different intrinsics;
and~$\alpha>0$ is the weighting between the task duration
and the control performance.
The constraints require that:
the object should reach the goal position;
each robot can only participate in the pushing of one object at any time;
the feasibility conditions
detailed in Appendix~\ref{app:model} should hold;
and all objects and robots should be collision-free.

Note that the design of function~$\mathrm{J}_m(\cdot)$ is non-trivial
and technically involved; details are given in the sequel.
Moreover, the CHO problem in~\eqref{eq:problem} has an extremely large decision space:
there are $M^{N}$ robot-object allocations per decision epoch and
exponentially many timed schedules of pushing modes and contact forces over the horizon.
Even ignoring discrete choices, the continuous variables already scale as
$\approx (3N+3M)\,\overline{T}$ when optimizing all objects jointly,
rendering the overall feasible set high-dimensional and highly nonconvex,
where $\overline{T}$ denotes the task duration.

%% file: contents/solution.tex
\section{Proposed Solution}\label{sec:solution}
The proposed solution consists of three interleaved components
in the top-down structure,
which are triggered at different conditions and rates,
as shown in Fig.~\ref{fig:overall}.
In particular, the overall task of multi-object pushing is first decomposed
into numerous subtasks via MAPF with ordering constraints in Sec.~\ref{subsec:task},
which are assigned dynamically to subgroups of robots in a receding-horizon fashion.
Then, for each subgroup to execute the assigned subtask,
the optimal hybrid plan is generated via a hybrid search algorithm in Sec.~\ref{subsec:hybrid},
which is accelerated by a diffusion-based predictor
for keyframes and pushing modes.
Finally, to cope with uncertainties during execution,
the object trajectory, the hybrid plan and the task assignment
are all updated online, as described in Sec.~\ref{subsec:online}.
Discussions are provided in Sec.~\ref{subsec:discuss}.

\input{contents/task.tex}

\input{contents/hybrid.tex}

\input{contents/online.tex}

\input{contents/general.tex}

%% file: contents/task.tex

\subsection{Decomposition and Assignment of Pushing Tasks}\label{subsec:task}
Since the desired goal state of each object can be far away
and the objects might be blocking each other,
the multi-agent path finding (MAPF) algorithm
is adopted first to synthesize a timed path for each object to reach its goal states.
Although these paths are spatially and temporally collision-free,
the object velocity during collaborative pushing is hard to predict and control, 
making it difficult to follow the paths accurately in time and space.
Thus, these paths are further decomposed into smaller segments with
temporal ordering, as subtasks of the overall pushing task.
Lastly, to cope with the uncertainties during execution,
a dynamic task assignment algorithm is proposed to assign
the subtasks to subgroups of robots as coalitions online
in a receding-horizon fashion.

\begin{figure}[t!]
  \centering
  \includegraphics[width=0.95\linewidth]{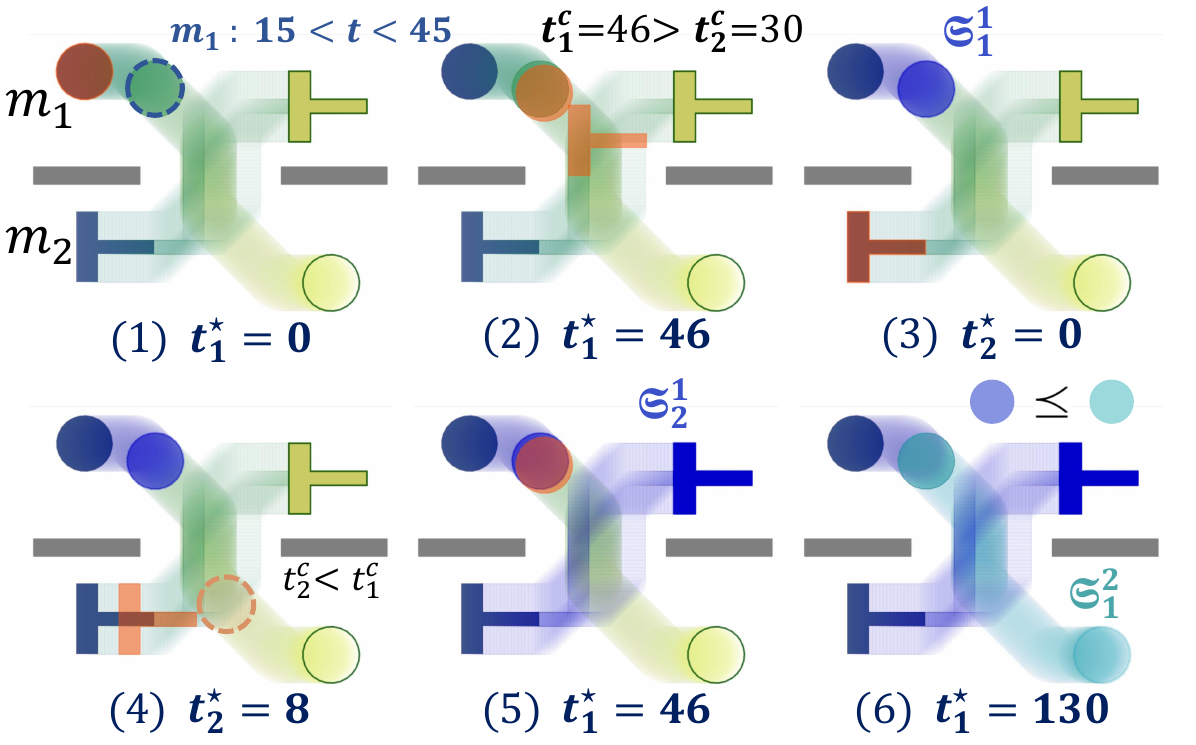}
  \hspace{-0.2in}
  \caption{
  {
  Illustration of the decomposition and ordering of pushing tasks
  via Alg.~\ref{alg:segments}, yielding $3$
  subtasks~$\{\mathfrak{S}^1_1,\mathfrak{S}^2_1,\mathfrak{S}^1_2\}$
  with ordering $\mathfrak{S}^1_2\preceq \mathfrak{S}^2_1$
  and $\mathfrak{S}^1_1 \preceq \mathfrak{S}^2_1$.
  Note that object~$m_1$ waits for object~$m_2$ to pass the corridor.
  }}
  \label{fig:segmentation}
  \vspace{-5mm}
\end{figure}

\subsubsection{Decomposition and Ordering via MAPF}\label{subsubsec:mapf}
Given the initial and goal states~$(\mathbf{s}^{\texttt{0}}_m,
\mathbf{s}^{\texttt{G}}_m)$ of each object~$m\in \mathcal{M}$,
a MAPF algorithm~\cite{stern2019multi} is employed to generate a set of
timed path for all objects:
$\widehat{\mathfrak{S}} \triangleq
  \{\mathfrak{S}_m,\forall m\in \mathcal{M}\}$,
where each path~$\mathfrak{S}_m$ of length~$L>0$
for object~$m$ is defined as:
\begin{equation}\label{eq:m-path}
\mathfrak{S}_m\triangleq (t_0,\,\mathbf{s}_m(t_0))
\cdots (t_\ell,\,\mathbf{s}_m(t_\ell))
\cdots (t_L,\,\mathbf{s}_m(t_L)),
\end{equation}
with~$\{t_\ell\}$ being the uniform time steps
and~$\mathbf{s}_m(t_\ell)$ being the object state at time~$t_\ell$.
Note that MAPF is a centralized algorithm which ensures that
the objects do not collide with each other nor the obstacles along the timed paths,
i.e.,
$\Omega_{m_1}(t_\ell)\cap \Omega_{m_2}(t_\ell)=\emptyset, \forall m_1\neq m_2$, $\forall t_\ell$.
Due to the combinatorial complexity w.r.t. the number of robots
and the collision detection of non-convex polytopes,
a sequential planning approach can also be adopted
to reduce computation time, e.g.,
ordered by the length of shortest path to the goal states.
Although the paths in~$\widehat{\mathfrak{S}}$ derived above are collision-free,
they can be difficult to follow both spatially and temporally.
Thus, to ensure safety during execution,
these paths are further decomposed into smaller segments
that are temporally ordered.
For ease of notation, let $\mathfrak{S}_m(t^\texttt{s},t^\texttt{e})$
denote the segment of~$\mathfrak{S}_m$
between the time period~$[t^\texttt{s},\, t^\texttt{e}]$,
i.e.,
$\mathfrak{S}_m \triangleq
(t^\texttt{s},\,\mathbf{s}_m(t^\texttt{s})) \cdots
(t^\texttt{e},\,\mathbf{s}_m(t^\texttt{e}))$.
Then, each path can be decomposed into a sequence of segments.

\begin{definition}[Path Segments]\label{def:segments}
Each path~$\mathfrak{S}_m\in \widehat{\mathfrak{S}}$ can be decomposed
into~$K_m\geq 1$ segments
according to its spatial intersection with other paths, i.e.,
\begin{equation}\label{eq:m-seg}
\mathfrak{S}_m \triangleq \mathfrak{S}^1_m \cup
\cdots \mathfrak{S}^{k}_m \cup \cdots \mathfrak{S}^{K_m}_m,
\end{equation}
where~$\mathfrak{S}^{k}_m\triangleq \mathfrak{S}_m(t^{k,\texttt{s}}_m,t^{k,\texttt{e}}_m)$
is the~$k$-th segment of~$\mathfrak{S}_m$
between the starting time~$t^{k,\texttt{s}}_m$
and the ending time~$t^{k,\texttt{e}}_m$,
where~$
0\leq t^{k,\texttt{s}}_m<t^{k,\texttt{e}}_m=
t^{k+1,\texttt{s}}_m < t^{k+1,\texttt{e}}_m \leq L$.
\hfill $\blacksquare$
\end{definition}

\begin{algorithm}[t!]
  \caption{Segmentation and Partial Ordering}
  \label{alg:segments}
  \SetAlgoLined
  \KwIn{Paths $\{\mathfrak{S}_m\}$}
  \KwOut{Segments $\{\overline{\mathfrak{S}}_m\}$}
  Initialize $t^\star_m= 0,\overline{\mathfrak{S}}_m=\emptyset\ \forall m\in \mathcal{M}$;\\
  \While{$\exists m \in \mathcal{M}, t_{m}^\star< t_L$}{
      \For{$m\in\mathcal{M}$ and $t^\star_m< t_L$}{
      Determine the next instance~$t^\texttt{s}_m$ by~\eqref{eq:split};\\
      \textbf{if}{\ $t^\texttt{s}_m$ not found\ }\textbf{then}{\ $t^\texttt{s}_m$ set to $t_L$;}\\
      \textbf{if}{\ $t^\texttt{s}_m = t^\star_m$\ }\textbf{then}{\ \textbf{Continue};}\\
      Add $\mathfrak{S}_m(t^\star_m,t^{\texttt{s}}_m)$ to $\overline{\mathfrak{S}}_m$;\\
      $t^\star_m\leftarrow t^{\texttt{s}}_m$;
      }
  }
  Determine the partial relations for $\{\overline{\mathfrak{S}}_m\}$ by Def.~\ref{def:ordering};\\
\end{algorithm}

Thus, denote by~$\overline{\mathfrak{S}}_{m}\triangleq \{\mathfrak{S}^{k}_m,\,k=1,\cdots K_m\}$
the set of segments for each object,
and~$\overline{\mathfrak{S}} \triangleq \{\overline{\mathfrak{S}}_m,\,\forall m\in \mathcal{M}\}$
for all objects, respectively.
In addition, the cumulative covered area of each segment~$\mathfrak{S}^{k}_m$
is defined as:
$S^{k}_m\triangleq \Omega_m(t^{k,\texttt{s}}_m)\cup \cdots \cup \Omega_m(t^{k,\texttt{e}}_m)$,
where~$\Omega_m(t)$ is the covered area of object~$m$ at~$\mathbf{s}_m(t)$.
More importantly, the segments are temporally ordered
by a partial relation defined as follows.
\begin{definition}[Partial Ordering]\label{def:ordering}
The partial relation~$\preceq \subset
\overline{\mathfrak{S}} \times \overline{\mathfrak{S}}$
satisfies that~$\mathfrak{S}^{k_1}_{m_1} \preceq \mathfrak{S}^{k_2}_{m_2}$
if one of the following cases hold:
(I)~$m_1=m_2$ and $k_1< k_2$;
(II)~$m_1\neq m_2$
and $t^{k_1,\texttt{c}}_{m_1} \leq t^{k_2,\texttt{c}}_{m_2}$ holds,
where
$t^{k_1,\texttt{c}}_{m_1},t^{k_2,\texttt{c}}_{m_2}$
are the earliest time instances when the
segments~$\mathfrak{S}^{k_1}_{m_1}$
and~$\mathfrak{S}^{k_2}_{m_2}$ collide, i.e.,
\begin{equation}\label{eq:cross}
    t^{k_1,\texttt{c}}_{m_1}=
    \textbf{min}
    \Big{\{}
    t_\ell\in [t^{k,\texttt{s}}_{m_1},\, t^{k,\texttt{e}}_{m_1})\,|\,
      \Omega_{m_1}(t_\ell)\cap S^{k}_{m_2}\neq\emptyset
    \Big{\}},
\end{equation}
and~$t^{k_2,\texttt{c}}_{m_2}$ is defined analogously;
(III)~$m_1\neq m_2$, $S^{k_1}_{m_1}
\cap S^{k_2}_{m_2}=\emptyset$,
and there exists~$\mathfrak{S}^{k}_{m}\in \overline{\mathfrak{S}}\setminus
\{{\mathfrak{S}^{k_1}_{m_1},\mathfrak{S}^{k_2}_{m_2}}\}$
such that
$\mathfrak{S}^{k_1}_{m_1} \preceq \mathfrak{S}^{k}_{m}
\preceq\mathfrak{S}^{k_2}_{m_2}$ holds. \hfill $\blacksquare$
\end{definition}

Namely, all segments of the same object are ordered
by their temporal sequence,
while the segments of different objects
that can potentially intersect are ordered by the time that intersection first occurs.
Denote
by~$\texttt{Pre}(\mathfrak{S}^{k_2}_{m_2})
\triangleq \{\mathfrak{S}^{k_1}_{m_1}\in \overline{\mathfrak{S}}
\,|\,\mathfrak{S}^{k_1}_{m_1} \preceq \mathfrak{S}^{k_2}_{m_2} \}$
these subtasks that are ordered before~$\mathfrak{S}^{k_2}_{m_2}$.

\begin{problem}\label{prob:decompose}
Given the paths~$\{\mathfrak{S}_m\}$,
find the segments~$\{\overline{\mathfrak{S}}_m\}$
such that the partial ordering~$\preceq$  above is satisfied.
\hfill $\blacksquare$
\end{problem}

{As summarized in Alg.~\ref{alg:segments}
and shown in Fig.~\ref{fig:segmentation},
an iterative algorithm is proposed to derive the segments~$\overline{\mathfrak{S}}$
from the paths~$\widehat{\mathfrak{S}}$.}
For each object~$m$,
the variable~$t^\star_m$ is introduced
to track the largest splitting time instance,
which is set to zero initially.
Then, the next splitting instance~$t^{\texttt{s}}_m$ is given by:
\begin{equation}\label{eq:split}
      t^{\texttt{s}}_m\triangleq
      \underset{m'\neq m}{\textbf{min}}
      \Big{\{}t^\texttt{c}_m\,
      |\, t^\texttt{c}_m\,>t^\texttt{c}_{m'} \text{ by~\eqref{eq:cross} given}~\widetilde{\mathfrak{S}}_m,\widetilde{\mathfrak{S}}_{m'}
      \Big{\}},
\end{equation}
where~$\widetilde{\mathfrak{S}}_m\triangleq
\mathfrak{S}_m(t^\star_m,t_L)$
and $\widetilde{\mathfrak{S}}_{m'} \triangleq
\mathfrak{S}_{m'}(t^\star_{m'},t_L)$
are the remaining paths of objects~$m$ and~$m'$ that have not been segmented.
If the splitting instances can not be found,
$t^{\texttt{s}}_{m}$ is set to $t_L$.
Otherwise, a new segment~$\mathfrak{S}_m(t^\star_m,t^\texttt{s}_m)$
is generated and~$t^\star_m$ is updated to $t^\texttt{s}_m$.
This process is repeated until~$t^\star_m=t_L$ holds, $\forall m\in\mathcal{M}$.
Lastly, the partial ordering among the segments is determined by the
rules in Def.~\ref{def:ordering}.

The above strategy of decomposition not only divides
the overall pushing task of an object into subtasks,
but also ensures a collision-free execution under uncertain
task durations,
as long as the partial ordering among the subtasks is satisfied.

\subsubsection{Dynamic Task Assignment}\label{subsubsec:assign}
The set of segments~$\overline{\mathfrak{S}}$ above represents
the subtasks of the overall pushing task,
each of which could be accomplished by a subgroup of robots.
In other words, each robot would execute a sequence of subtasks,
i.e., push different objects along different segments.

\begin{definition}[Task plan]\label{def:plan}
The local task plan of robot~$i\in \mathcal{N}$ is denoted
by~$\boldsymbol{\tau}_i\triangleq (t_1,\,\mathfrak{S}^{k_1}_{m_1})
(t_2,\,\mathfrak{S}^{k_2}_{m_2})\cdots (t_{L_i},\,\mathfrak{S}^{k_{L_i}}_{m_{L_i}})$,
where the segment~$\mathfrak{S}^{k_\ell}_{m_\ell}\in \overline{\mathfrak{S}}$
is executed by robot~$i$ from time~$t_\ell$;
and~$L_i>0$ is the total length.
The overall task plan of the fleet is denoted by~$\overline{\boldsymbol{\tau}}
\triangleq \{\boldsymbol{\tau}_i,\, i\in \mathcal{N}\}$.
\hfill $\blacksquare$
\end{definition}

The overall task plan~$\overline{\boldsymbol{\tau}}$
is called \emph{valid} if the following two conditions hold:
(I) all partial orderings in~$\preceq$ are respected,
i.e., if~$\mathfrak{S}^{k_1}_{m_1} \preceq \mathfrak{S}^{k_2}_{m_2}$,
then the segment~$\mathfrak{S}^{k_1}_{m_1}$ should be completed
before the segment~$\mathfrak{S}^{k_2}_{m_2}$
is started by the assigned subgroup of robots.
(II) the subgroup of robots assigned to each segment
should be sufficient to push the object along the segment.
(III) there should be enough time for each robot~$i\in \mathcal{N}$ to
navigate between consecutive segments in its plan.

\begin{problem}\label{prob:assign}
The objective of the task assignment for~$\overline{\mathfrak{S}}$
is to find a valid task plan
such that the estimated time to complete all tasks is minimized, i.e.,
$\underset{\overline{\boldsymbol{\tau}}}{\textbf{min}}
\Big\{\underset{\mathfrak{S}^{k}_{m}\in \overline{\mathfrak{S}}}{\textbf{max}}
\big{\{}t_{\texttt{end}}(\mathfrak{S}^{k}_{m},\,
\mathcal{N}^{k}_{m})\big{\}}\Big\}$.
\hfill $\blacksquare$
\end{problem}

\begin{figure}[t!]
  \centering
  \includegraphics[width=0.95\linewidth]{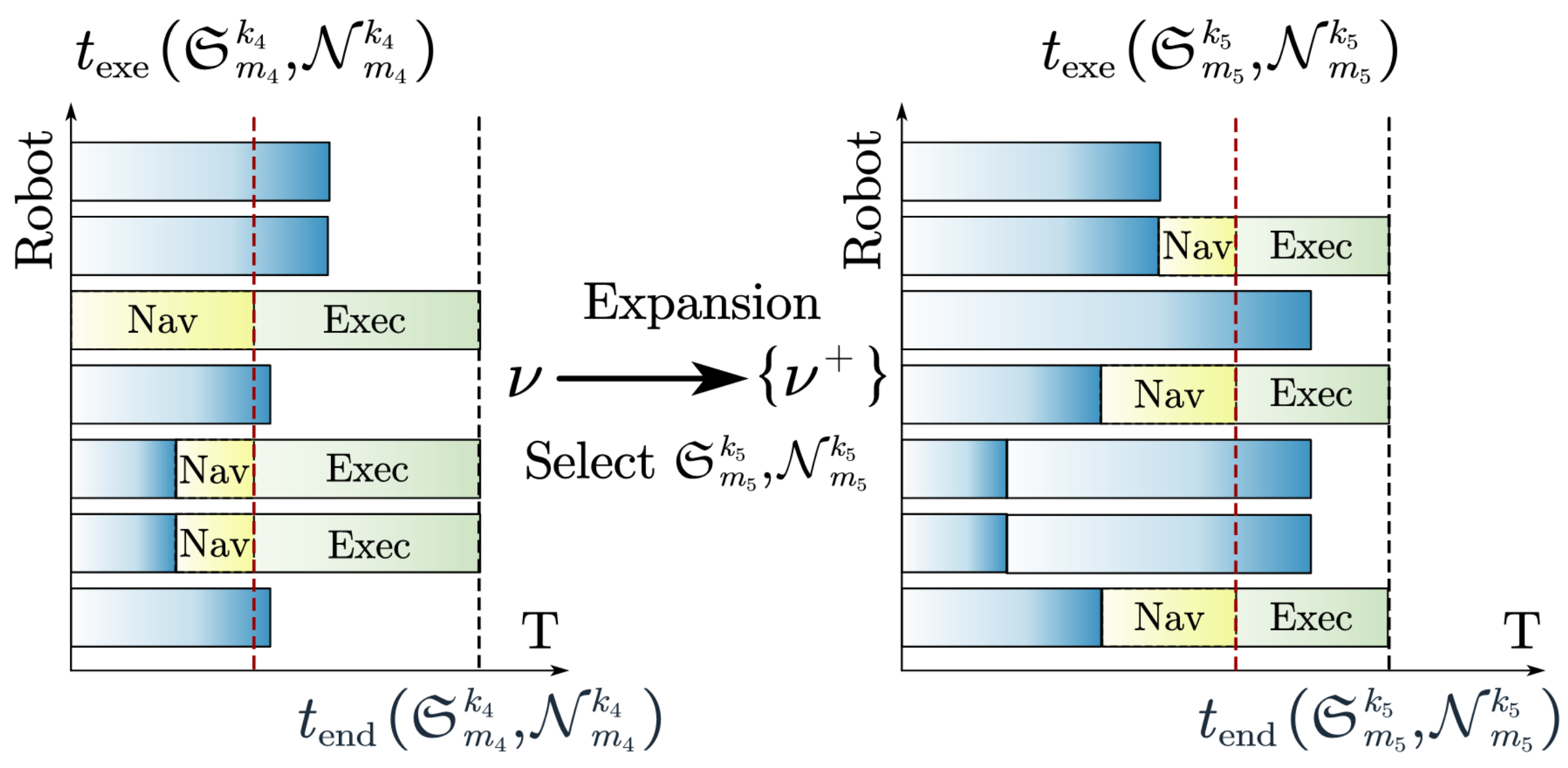}
  \vspace{-0.1in}
  \caption{{Selection and expansion during
    the proposed receding-horizon assignment of the
    {12 subtasks and 7 robots},
    given the current planning time (red dashed line)
    and the horizon~$H=5$ (blue dashed line).}}
  \label{fig:assign}
  \vspace{-4mm}
\end{figure}
\begin{figure*}[t!]
  \centering
  \includegraphics[width=0.9\linewidth]{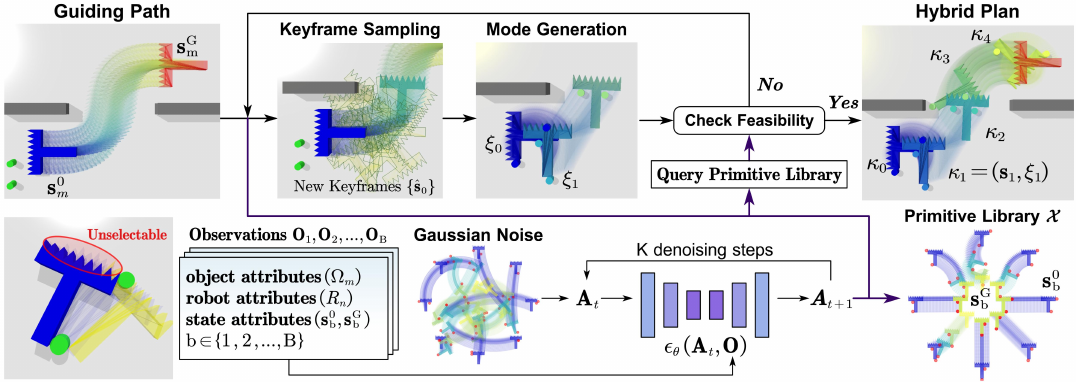}
  \vspace{-0.1in}
  \caption{
  {Illustration of the keyframe-guided hybrid search algorithm (\textbf{Top}),
   which is accelerated by the diffusion-based predictor for keyframes and pushing modes (\textbf{Bottom}).
  Note that the multi-modal predictions are verified by the hybrid search scheme for feasibility and quality.}}
  \label{fig:neural}
  \vspace{-4mm}
\end{figure*}

{
  To begin with, the above problem includes the job-shop problem
  as a special instance~\cite{torreno2017cooperative},
  thus is also NP-hard. The most straightforward solution
is to formulate a Mixed Integer Linear Program (MILP).
However, the computation complexity and the lack of intermediate solutions
make it unsuitable for this application,
particularly so when the pushing subtasks have large uncertainties in execution time.}

{Thus,
an adaptive receding-horizon task assignment algorithm is proposed,
which incrementally assigns subtasks in~$\overline{\mathfrak{S}}$
to robot subgroups via node expansion.
The details are omitted here due to limited space.
Briefly speaking, it processes at most~$H>0$ subtasks per planning
horizon to avoid intractable complexity
 and account for uncertainties during collaborative pushing.
As illustrated in Fig.~\ref{fig:assign},
the search begins with an empty assignment node,
and expands it by assigning one subtask at a time,
ensuring that preceding subtasks are assigned first
as required by the ordering in~$\preceq$.
Each robot subgroup assigned to a subtask is evaluated by
the feasibility and the estimated completion time,
which is described in the subsequent module.
The nodes are prioritized based on their time-average efficiency.
The search is terminated after all~$H$ subtasks are assigned.
Thus, the node with the maximum efficiency
is selected as the partial plan~$\overline{\boldsymbol{\tau}}$,
by which the subtasks are executed.
Once the replaning condition holds such as a fixed number of
subtasks are fulfilled or certain subtasks are delayed,
the set of subtasks~$\overline{\mathfrak{S}}$ are updated
and re-assigned.
}

{More details about the dynamic task assignment
  can be found in the supplementary materials.
  Moreover, the above procedure for task decomposition and partial ordering
  have the following guarantees for completeness and correctness,
of which the proofs are provided in the Appendix~\ref{app:proof}.}

\begin{lemma}\label{lemma:decompose}
    Alg.~\ref{alg:segments} terminates in finite steps
    and generates a strict partial ordering.
    \hfill $\blacksquare$
  \end{lemma}

  \begin{lemma}\label{lemma:ordered_execution}
  Given the partially-ordered segments~$(\overline{\mathfrak{S}},\, \preceq)$,
  each object~$m\in \mathcal{M}$ can reach its goal state without collision,
  by traversing its sequence of segments~$\overline{\mathfrak{S}}_m$
  as follows:
  (I) each segment~$\overline{\mathfrak{S}}^k_m\in \overline{\mathfrak{S}}_m$
  takes a bounded time to traverse;
  (II) if~$\mathfrak{S}^{k_1}_{m_1} \preceq \mathfrak{S}^{k_2}_{m_2}$ holds,
  object~$m_2$ can be moved from the initial
  state~$\mathbf{s}_{m_2}(t^{k_2,\texttt{s}}_{m_2})$
  of~$\mathfrak{S}^{k_2}_{m_2}$
  \emph{only} after object~$m_1$ has reached the ending
  state~$\mathbf{s}_{m_1}(t^{k_1,\texttt{e}}_{m_1})$
  of~$\mathfrak{S}^{k_1}_{m_1}$,
  $\forall \mathfrak{S}^{k_2}_{m_2}\in \texttt{Pre}(\mathfrak{S}^{k_1}_{m_1})$.
  \hfill $\blacksquare$
  \end{lemma}

%% file: contents/hybrid.tex
\def\sss{\scriptscriptstyle{}}
\def\bsss{\sss{\mathrm{B}}}

\subsection{Accelerated Hybrid Optimization for Collaborative Push}\label{subsec:hybrid}
Given the local task plans~$\overline{\boldsymbol{\tau}}=\{\boldsymbol{\tau}^\star_i\}$
from the previous section,
a subgroup of robots~$\mathcal{N}_m$ is assigned to push object~$\Omega_m$ along the path segment~$\mathfrak{S}^k_m$.
This section describes how to efficiently
generate feasible hybrid plans for the subgroup
as a sequence of pushing modes, forces and the reference trajectory,
in a learning-while-planning manner.

\begin{problem}\label{prob:assign}
Given~$\mathfrak{S}^k_m$ and $\mathcal{N}^k_m$ for object~$\Omega_m$,
determine the hybrid plan~$\{(\boldsymbol{\xi}(t),\mathbf{S}_{\mathcal{N}_m}(t),\mathbf{s}_m(t))\}$
such that the local cost function in~\eqref{eq:problem} is minimized.
\hfill $\blacksquare$
\end{problem}

\subsubsection{Keyframe-guided Hybrid Search}\label{subsubsec:hybrid-search}

A keyframe–guided hybrid search (KGHS) is proposed to solve
Problem~\ref{prob:assign} as shown in Fig.~\ref{fig:neural}. Instead of optimizing an
entire segment at once, the algorithm recursively splits it into short
\emph{arc segments}, each of which is executable under a \emph{single} pushing mode.
A \emph{keyframe} is an intermediate object state at which
the contact mode is allowed to switch.
Let \(\mathbf{s}_\ell\) denote the \(\ell\)-th keyframe along the segment.
Namely, the hybrid plan becomes a sequence of keyframes and the pushing modes between them,
denoted by~$\vartheta\triangleq \kappa_0\cdots\kappa_\ell \cdots \kappa_{L_{\vartheta}}$,
with the~$\ell$-th stage~$\kappa_\ell\triangleq (\mathbf{s}_\ell, \boldsymbol{\xi}_\ell)$ being
the keyframe~$\mathbf{s}_\ell$ and the mode~$\boldsymbol{\xi}_\ell=(\xi_\ell,\mathbf{F}_\ell)$,
$\forall \ell=1,\cdots,L_{\vartheta}$.
Note that~$\mathbf{s}_0=\mathbf{s}^\texttt{0}_m$ and $\mathbf{s}_{L_{\vartheta}}=\mathbf{s}^\texttt{G}_m$,
while~$\boldsymbol{\xi}_\ell$ is the pushing mode for the arc segment~$\varrho_\ell
\triangleq \overgroup{\mathbf{s}_\ell \mathbf{s}_{\ell+1}}$.
The search space, denoted by~$\Theta\triangleq \{\vartheta\}$, is thus significantly reduced.

Starting from~$\vartheta_0\triangleq (\mathbf{s}^\texttt{0}_m,\emptyset)(\mathbf{s}^\texttt{G}_m,\emptyset)$,
the search space~$\Theta$ is explored via iteratively selecting the promising node and adding keyframes as needed.
To begin with, the node $\vartheta^\star$ with the lowest estimated cost is selected, i.e.,
\begin{equation}\label{eq:est-cost}
  \begin{split}
  \mathrm{J}^m_{\texttt{hyb}}(\vartheta)\triangleq
  \sum_{l=0}^{L_{\vartheta}-1} \big{(}&\mathrm{J}^m_{\texttt{MF}}(\boldsymbol{\xi}_{\ell},\, \mathbf{p}^{\bsss}_{\varrho_\ell})
  +w_{\texttt{s}}\mathrm{J}^m_{\texttt{sw}}(\boldsymbol{\xi}_\ell,\,\boldsymbol{\xi}_{\ell+1})\\
  &+w_{\texttt{n}}\mathrm{J}^m_{\texttt{nv}}(\mathbf{s}_{\ell},\, \mathbf{s}_{\ell+1})\big{)},
  \end{split}
\end{equation}
where~$\mathrm{J}^{m}_{\texttt{MF}}(\cdot)$
is a multi-directional estimation of the feasibility
of employing mode~$\boldsymbol{\xi}_{\ell}$ for the body-frame velocity $\mathbf{p}^{\bsss}_{\varrho_\ell}$
corresponding to the arc~$\varrho_\ell$,
with more detailed derivations provided in Appendix~\ref{app:feasibility};
$\mathrm{J}^m_{\texttt{sw}}(\cdot)$ measures the switching time from mode~$\boldsymbol{\xi}_\ell$
to mode~$\boldsymbol{\xi}_{\ell+1}$;
$\mathrm{J}^m_{\texttt{nv}}(\cdot)$ measures the navigation time cost along the arc $\varrho_\ell$;
and~$w_{\texttt{s}},w_{\texttt{n}}>0$ are the weighting parameters.
These cost terms are designed to reflect the objective function in Problem~\ref{eq:problem}.
Unlike a hard feasibility constraint~\cite{rosenfelder2024force}, the soft measure $\mathrm{J}^{m}_{\texttt{MF}}(\cdot)$
keeps the hybrid search \emph{solvable and robust} under modeling uncertainty or 
when the constraint is structurally unsatisfiable (e.g., single-contact pushing),
while still permitting
explicit enforcement of feasibility, as detailed in Appendix~\ref{app:feasibility}.
Afterwards, the selected node~$\vartheta^\star$ is expanded by finding the first
keyframe that has not been assigned with a mode,
e.g.,~$(\mathbf{s}_\ell,\emptyset)$ for the segment~$\varrho_\ell$.
Then, the node expansion is performed in four steps:

\begin{algorithm}[t!]
  \caption{Neural Accelerated Hybrid Optimization: $\texttt{HybDIF}(\cdot)$}
  \label{alg:hybrid}
  \SetAlgoLined
  \KwIn{Subgroup~$\mathcal{N}^k_m$,
        subtask~$\mathfrak{S}^k_m$,
        Library~$\mathcal{X}$}
  \KwOut{Hybrid Plan~$\vartheta_m^{k,\star}$,
    control~$\mathbf{u}_{\mathcal{N}^k_m}$.}
  \tcc{\textbf{Offline Generation}}
  {Collect state pairs $\{(\mathbf{s}_{\texttt{s}},\mathbf{s}_{\texttt{e}})\}$
  along the segment~$\mathfrak{S}^k_m$};\\
  Batch-generate hybrid plans for all state pairs:
  $\{\widehat{\vartheta}^{\star}\}\leftarrow \texttt{DIF}(\Omega_m,\mathcal{N}_m^k,\{(\mathbf{s}_{\texttt{s}},\mathbf{s}_{\texttt{e}})\})$;\\
  Update library $\mathcal{X} \leftarrow \mathcal{X} \cup \{\widehat{\vartheta}^{\star}\}$;\\

  \tcc{\textbf{Keyframe-guided Hybrid Search}}
  Initialize~$\Theta=\{\vartheta_0\}$;\\
  \While{not terminated}
  {
    Select~$\vartheta^\star$ by~\eqref{eq:est-cost};\\
    Find first keyframe~$(\mathbf{s}_\ell,\emptyset)\in \vartheta^\star$ without a mode;\\
    \If{Arc~$\varrho_\ell$ has collision}
    {
      $\widehat{\vartheta}^{\star}
      \leftarrow
      (\mathbf{s}_{\ell},\emptyset)(\widehat{\mathbf{s}}_{\ell},\emptyset),
      \widehat{\mathbf{s}}_{\ell}\in\mathfrak{S}^k_m$;\\
    }
    \Else
    {
      \tcc{\textbf{\footnotesize{Sequentially run the following steps
      until $\widehat{\vartheta}^{\star}$ becomes feasible
      }}}
      $\widehat{\vartheta}^{\star}\leftarrow(\widehat{\kappa}^{\star}_1,\cdots,\widehat{\kappa}^{\star}_h)
      =\mathcal{X}(\Omega_m,\mathcal{N}^k_m,\mathbf{s}_{\ell},\mathbf{s}_{\ell+1})$;\\
      $\widehat{\vartheta}^{\star}\leftarrow
      \texttt{DIF}(\Omega_m,\mathcal{N}_m^k,\mathbf{s}_\ell,\mathbf{s}_{\ell+1})$;\\
      { $\widehat{\vartheta}^{\star}\leftarrow \texttt{IterSamp}(\vartheta^\star,\varrho_\ell)$ };\\
      $\widehat{\vartheta}^{\star}\leftarrow
      (\mathbf{s}_{\ell},\emptyset)(\widehat{\mathbf{s}}_{\ell},\emptyset),
      \, \widehat{\mathbf{s}}_{\ell}\in\mathfrak{S}^k_m$;\\
      {
        \If{$\|\varrho_{\ell}\|<\epsilon$}
        {
          $\widehat{\vartheta}^{\star}\leftarrow \texttt{SeqArcApprox}(\varrho_\ell,\mathcal{N}^k_m)$;\\
        }
      }
    }
    Expand~$\vartheta^\star$ by replace
    $(\mathbf{s}_\ell,\emptyset)$ with $\widehat{\vartheta}^{\star}$;
  }
\end{algorithm}

(I) If the arc~$\varrho_\ell$
intersects with an obstacle, a new keyframe~$(\widehat{\mathbf{s}}_{\ell},\,\emptyset)$
is inserted between~$\mathbf{s}_\ell$ and $\mathbf{s}_{\ell+1}$ by selecting an
intermediate collision-free state~$\widehat{\mathbf{s}}_{\ell}$, i.e., $\vartheta^+ \triangleq
\cdots \kappa_\ell (\widehat{\mathbf{s}}_{\ell},\emptyset)\kappa_{\ell+1}\cdots$;
(II) If the arc~$\varrho_\ell$ is collision-free,
a mode is generated for the segment~$\varrho_\ell$
by minimizing the loss
function~$J^m_{\texttt{MF}}(\cdot)$ in~\eqref{eq:est-cost} to measure feasibility,
i.e.,
\begin{equation}\label{eq:mode_gen}
  \boldsymbol{\xi}^\star_\ell = \textbf{argmin}_{\boldsymbol{\xi}_\ell\in\Xi_m} \big{\{}
  \mathrm{J}^m_{\texttt{MF}}(\boldsymbol{\xi}_\ell,\, \mathbf{p}^{\bsss}_{\varrho_\ell})\big{\}},
\end{equation}
which results in a combinatorial and nonlinear optimization problem.
To solve~\eqref{eq:mode_gen}, we employ a parallel multi-start search over contact modes.
Candidate modes are initialized by random sampling and iteratively refined by adjusting contact points.
For each candidate $\boldsymbol{\xi}_\ell$, we first check \emph{force} feasibility via a loss related to force balance, 
and then \emph{practical} feasibility by simulating the pushing process along $\varrho_\ell$.
A mode is accepted only if both criteria are satisfied; see Appendix~\ref{app:feasibility} for details.
Thus, if a feasible mode is found, the mode is assigned to the segment~$\varrho_\ell$.

(III) If no feasible mode is found between $\mathbf{s}_\ell$ and $\mathbf{s}_{\ell+1}$,
an iterative sampling procedure is proposed to generate
a sequence of keyframes and modes
that further split the segment~$\varrho_\ell$ into~$h>0$
sub-segments, i.e.,
\begin{equation}\label{eq:iter-sample}
  \widehat{\vartheta}^{\star}
  \triangleq (\widehat{\kappa}^{\star}_1,\ldots,\widehat{\kappa}^{\star}_h)
  \triangleq \texttt{IterSamp}(\mathbf{s}_{\ell},\,\mathbf{s}_{\ell+1}). 
\end{equation}
Keyframes are initially sampled uniformly along $\varrho_\ell$ and then perturbed with diminishing variance; 
for each perturbed segment, modes are generated as in (II) and evaluated using~\eqref{eq:est-cost}.
The procedure terminates when all sub-segments admit practically feasible modes or a maximum iteration count is reached.

(IV) Lastly, if the subplan $\widehat{\vartheta}^{\star}$ is still infeasible, 
a new keyframe $(\widehat{\mathbf{s}}_{\ell},\emptyset)$ is inserted 
between $\mathbf{s}_\ell$ and $\mathbf{s}_{\ell+1}$, thereby splitting $\varrho_\ell$ into two shorter segments.
When the arc length $\|\varrho_{\ell}\|$ is below a threshold $\epsilon>0$, 
we approximate the original arc by concatenating pre-validated modes,
\begin{equation}\label{eq:approx}
  \widehat{\vartheta}^{\star} \triangleq \texttt{SeqArcApprox}(\varrho_\ell,\,\mathcal{N}^k_m),
\end{equation}
which serves as a ``last resort'' hybrid plan for the subgroup~$\mathcal{N}^k_m$ due to its frequent mode switching.
The following lemma summarizes the theoretical guarantee of the hybrid search; 
the proof is given in Appendix~\ref{app:proof}.

\begin{lemma}\label{lemma:hybrid}
 {
      Given that the subgroup~$\mathcal{N}^k_m$ is {mode-sufficient}
      as defined in Appendix.~\ref{app:sufficient},
      and the path~$\mathfrak{S}_m^k$ is collision-free,
      the proposed hybrid search scheme finds a feasible hybrid
      plan~$\vartheta^\star$ in finite steps.
    }
\end{lemma}

\begin{figure*}[t!]
  \centering
  \includegraphics[width=0.9\linewidth]{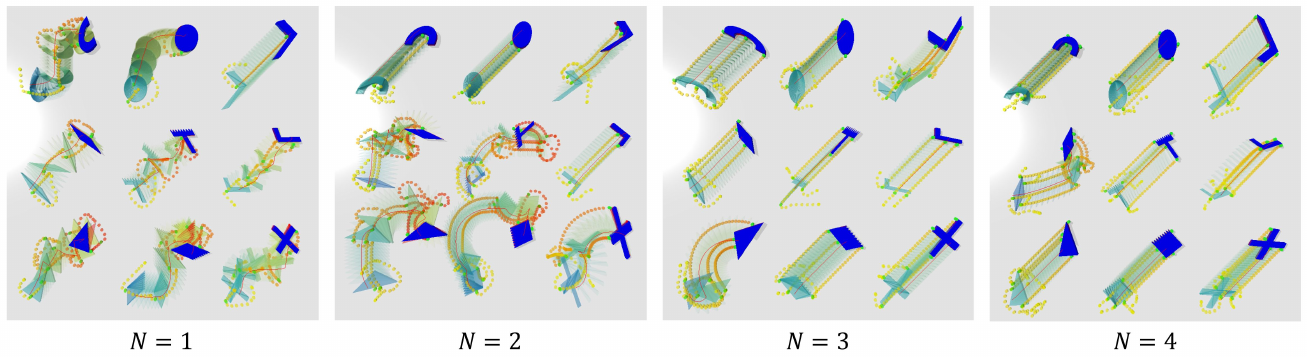}
  \vspace{-0.1in}
  \caption{{Dataset generation for training the diffusion-based
   generator of primitive pushing modes,
   where the initial state of the objects
   are randomized (in light green), the object shape is scaled
   and twisted,
   and the number of robots is varied from~$1$ to~$4$.
   The resulting robot trajectories are shown in red lines.}
 }
  \label{fig:data_gen}
  \vspace{-2mm}
\end{figure*}
\begin{figure*}[t!]
  \centering
  \includegraphics[width=0.9\linewidth]{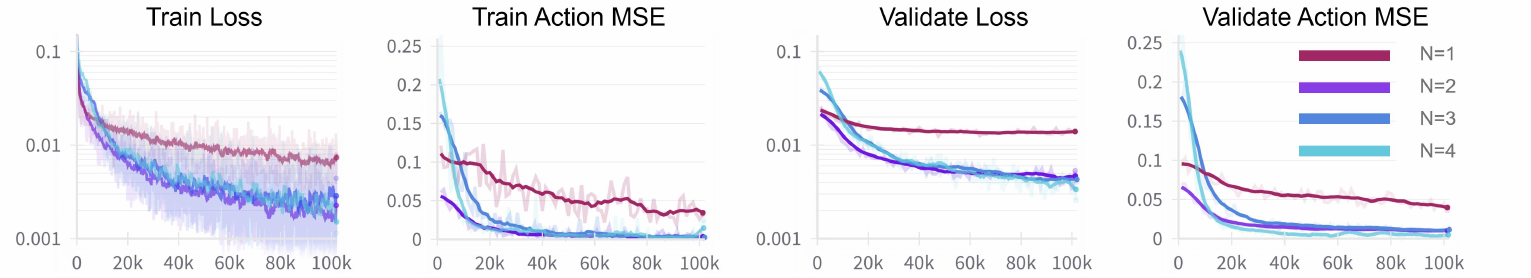}
  \vspace{-0.1in}
  \caption{{The training loss, the validation loss, the MSE
  of the hybrid plan during training and validation,
  for the pushing tasks shown in Fig.~\ref{fig:data_gen}.}}
  \label{fig:training}
  \vspace{-4mm}
\end{figure*}

\subsubsection{Diffusion-based Neural Acceleration}\label{subsubsec:diffusion}
A key aspect of the hybrid search scheme above is to generate suitable
keyframes~$\{\kappa_\ell\}$ along the path segments
and the associated pushing modes~$\{\boldsymbol{\xi}_\ell\}$.
To further improve the planning efficiency
and reduce planning time mainly for large-scale scenarios,
a diffusion-based neural accelerator
is proposed to generate plausible keyframes and pushing modes
for the hybrid search scheme.
Consequently, high-quality pushing strategies are found earlier
for any given path segment and even unseen objects,
in both free and cluttered environments.
In particular, it consists of three main components as described below:
the generation of training data,
the training of proposed network architecture,
and the deployment of the learned network.

\textbf{Dataset Generation}.
As shown in Fig.~\ref{fig:data_gen},
a large variety of pushing problems are instantiated
for randomly deformed basic shapes,
a random number of robots with random configurations,
and random initial and target states.
Then, the proposed keyframe-guided hybrid search scheme 
is used to solve each problem instance, 
yielding multiple candidate solutions.
Instead of terminating upon finding a feasible solution,
a fixed number of iterations is enforced to obtain better solutions.
Each candidate is evaluated in simulation, 
and the one with minimum task duration is selected.
For each selected solution, we store the problem description 
and its optimal hybrid plan:
\begin{equation}\label{eq:data}
  \mathfrak{D} \triangleq \Big\{\big(
  (\mathbf{s}_m^{\texttt{0}},\,\mathbf{s}_m^{\texttt{G}},
  \,\Omega_m,\,\{\mathbf{x}_n^{\texttt{0}}\}),\,
  (\vartheta^\star_m,\,T_m^\star)\big)\Big\}.
\end{equation}
To improve data efficiency, we also exploit two rotational symmetries.
We randomly rotate the object's body frame by an angle~$\theta$, 
expressing all contact points in the rotated frame and shifting the object orientation by~$-\theta$ (a pure change of reference), 
and use global in-plane rotational invariance to normalize each trajectory to a common zero terminal state,
thereby reducing the output degrees of freedom and letting the model predict in this canonical frame.

\textbf{Network Architecture and Training}.
The noise prediction network follows the U-net architecture
in~\cite{chi2023diffusionpolicy},
with action generation conditioned on observations~$\mathbf{O}$ via
Feature-wise Linear Modulation (FiLM)~\cite{ho2020denoising}.
The dataset~$\mathfrak{D}$ in~\eqref{eq:data}
is distilled into observational inputs~$\mathbf{O}$ and outputs~$\mathbf{A}$.
The observations~$\mathbf{O}$ include:
(I) object attributes, such as coordinates and normal vectors of a fixed number of candidate contact points 
and the maximum friction force and torque;
(II) robot attributes, including size and maximum pushing forces; and
(III) the initial and target object states $\mathbf{s}^{\texttt{0}}$ and $\mathbf{s}^{\texttt{G}}$.
The outputs $\mathbf{A}$ are sequences of keyframes and pushing modes,
\begin{equation}
  (\mathbf{s}_1,\boldsymbol{\xi}_1)\cdots(\mathbf{s}_h,\boldsymbol{\xi}_h)
  \leftarrow \texttt{DIF}(\Omega_m,\,\mathcal{N}_m^k,
  \,\mathbf{s}^{\texttt{0}},\,\mathbf{s}^{\texttt{G}}),
\end{equation}
where $\texttt{DIF}(\cdot)$ is the diffusion-based predictor.

\begin{figure*}[t!]
  \centering
  \includegraphics[width=0.9\linewidth]{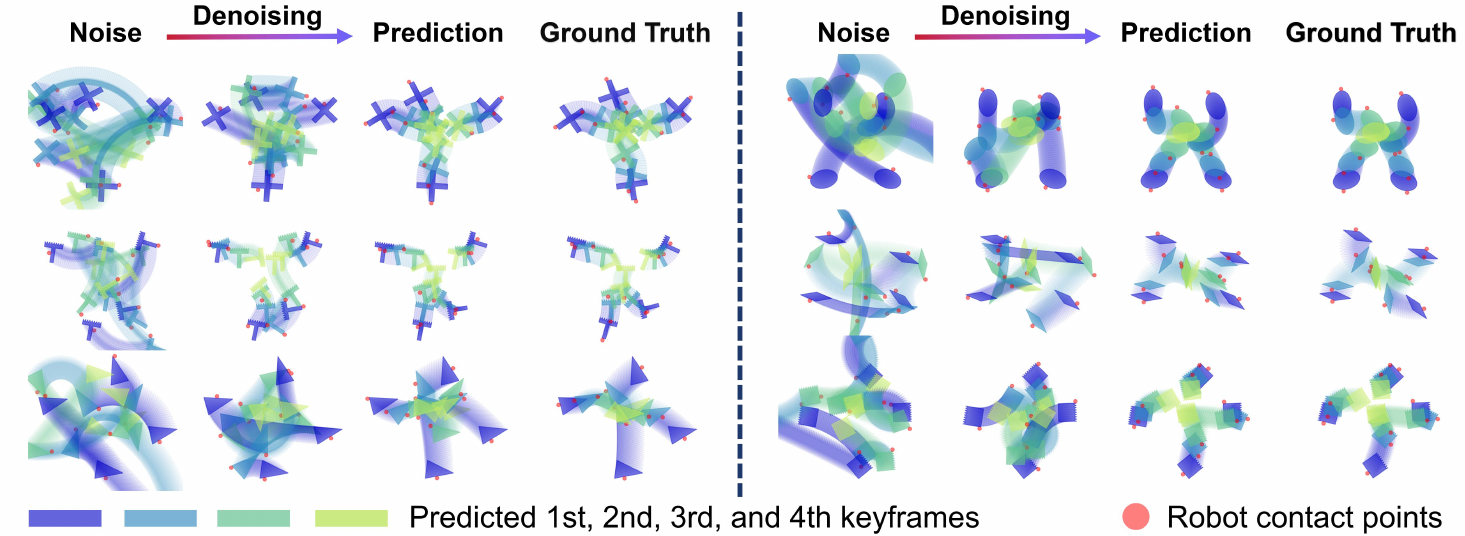}
  \vspace{-0.1in}
  \caption{
  {Illustration of generating hybrid plans with a horizon
  of~$3$ via the trained diffusion model,
  for $6$ different objects starting from~$4$
  distinctive initial positions and a common target
  at the origin.
  Each predicted hybrid plan consists of~$4$ keyframes
  and~$3$ pushing modes in between.
  During the denoising process,
  both the mode and keyframe quality improve progressively,
  with the final predictions align well with the ground truth in the dataset.}
  }
  \label{fig:primitive_diff}
  \vspace{-4mm}
\end{figure*}

The network is trained following the standard
Denoising Diffusion Probabilistic Model (DDPM) procedure~\cite{ho2020denoising}.
At a randomly sampled diffusion step $k$,
Gaussian noise $\epsilon^k$ is added to each action $\mathbf{a}\in\mathbf{A}$,
and the network is trained to predict $\epsilon^k$ from
$(\mathbf{o},\mathbf{a}+\epsilon^k,k)$ using the MSE loss
\begin{equation*}
  \mathcal{L}_{\boldsymbol \theta} \triangleq
  \mathrm{MSE}\big(\epsilon^k,\,
  \epsilon_{\boldsymbol \theta}(\mathbf{o},\mathbf{a}+\epsilon^k,k)\big),
\end{equation*}
where $\boldsymbol{\theta}$ denotes the network parameters.
As shown in Fig.~\ref{fig:training}, the training and validation losses,
together with the MSE of the predicted hybrid plans,
decrease steadily over epochs.
Fig.~\ref{fig:primitive_diff} further illustrates hybrid plans generated by the trained model for different objects and initial poses, 
where the keyframes and modes are progressively refined during denoising and align well with the ground-truth plans in the dataset.

\textbf{Deployment within Hybrid Search}.
The most straightforward way to deploy the learned neural network
is to replace the hybrid search and directly execute its predicted plans, 
or to replace the optimization-based mode generation in~\eqref{eq:mode_gen}.
However, this often leads to collisions and failures when test scenarios differ from the training data.
Instead, we use the diffusion model to propose \emph{initial} hybrid plans, 
which are then verified and, if necessary, locally refined by the model-based scheme.
Specifically, when the arc between $\mathbf{s}_{\ell}$ and $\mathbf{s}_{\ell+1}$ is collision-free, 
the diffusion model generates candidate keyframes and pushing modes connecting these states.
The best candidate is inserted into the search tree and then checked for feasibility and performance.
To further accelerate planning, we maintain an online library~$\mathcal{X}$ of previously verified pushing strategies, 
indexed by relative object displacements.
The library is queried first; if no suitable entry is found, the hybrid search with the diffusion model is invoked to generate a new verified strategy, 
which is then stored in~$\mathcal{X}$.
This ``learning while planning'' setup both speeds up solving a single problem and improves performance over multiple tasks.

\begin{remark}\label{remark:verification}
Unlike approaches that directly execute neural plans~\cite{kim2019learning,chi2023diffusionpolicy},
the predicted keyframes and modes are only prioritized within the hybrid search and are always \emph{verified}, 
which is crucial in unseen complex workspaces.
\hfill $\blacksquare$
\end{remark}

%% file: contents/online.tex
\subsection{Online Execution and Adaptation}\label{subsec:online}
Given the high-level assignment algorithm of the pushing tasks
and the hybrid optimization of the pushing policy,
the robotic fleet can start executing the pushing tasks
as follows.
\subsubsection{Mode Execution}\label{subsubsec:control}
Given the hybrid plan~$\vartheta_m^{k,\star}$ for each object~$m\in\mathcal{M}$
and subtask~$\mathfrak{S}^k_m$,
the robots need execute the plan by tracking the desired trajectories.
More specifically,
a local reference trajectory is generated
as the arc segment~$\varrho(t)$ that connects
the current state~${\mathbf{s}}_m(t)$ of the object
and the next keyframe state~${\mathbf{s}}^{\star}_m$ along the plan,
which is updated at each control loop.
Then, the next reference state~$\widehat{\mathbf{s}}^{\texttt{r}}_m$ is selected as the first state
along the planned arc~$\varrho(t)$ satisfying
$|\widehat{\mathbf{s}}^{\texttt{r}}_m - \mathbf{s}_m(t)| > \delta_{\texttt{c}}$, where~$\delta_{\texttt{c}} > 0$
is a fixed threshold.
Given this reference object state,
the corresponding contact point and orientation for each robot~$n \in \mathcal{N}^k_m$
as specified by the hybrid plan are denoted by~$\widehat{\mathbf{c}}_n$ and~$\widehat{\psi}_n$,
respectively,
yielding the desired robot state~$\widehat{\mathbf{s}}^{\texttt{r}}_n \triangleq (\widehat{\mathbf{c}}_n, \widehat{\psi}_n)$.
Then, the desired translational and rotational velocities are given by:
\begin{equation}\label{eq:control}
\widehat{\mathbf{v}}_n = K_{\texttt{vel}} \left(\widehat{\mathbf{c}}_n - \mathbf{c}_n\right),\quad
\widehat{\omega}_n = K_{\texttt{rot}} \left(\widehat{\psi}_n - \psi_n\right);
\end{equation}
where~$K_{\texttt{vel}}$ and~$K_{\texttt{rot}}$ are positive gains.
These velocity commands are then passed to the low-level controllers,
yielding actuation forces~$\mathbf{Q}_{\texttt{drv}}$ for robot~$R_n$.
The exact implementation of the low-level controllers depends on the actuation hardware.
Moreover, to handle contact slippage and deformation,
pushing is paused when any contact point deviates beyond a threshold,
triggering load-free motion to re-establish valid contact.
Lastly, for high precision tracking at centimeter level,
a more deliberate control scheme might be required,
e.g., distributed coordination and control~\cite{rosenfelder2024force,ebel2024cooperative}.
Note that $\mathbf{u}_n$ in~\eqref{eq:control} updates in real-time,
relying solely on the online computation of
the analytical arc segment $\varrho_t$ at each time step.

\begin{remark}\label{rm:control}
  {
  As neither the simulated nor physical robots in our work
  are equipped with force sensors,
  the kinematic control in~\eqref{eq:control}
  are adopted instead of explicit force-based control.
  Thus, contact forces emerge implicitly from physical interactions,
  reducing the reliance on accurate force measurements.
  More analyses on the feasibility
  of this kinematic control scheme
  can be found in Appendix~\ref{app:feasibility}.
  }
  \hfill $\blacksquare$
\end{remark}
\begin{remark}
  {
    Although the controller above is designed for holonomic models,
    extensions to non-holonomic robots such as
    quadruped and differential-drive robots
    can be achieved by local adjusting schemes,
    e.g., turn-and-forward.
    More numerical examples are given in Sec.~\ref{subsec:generalization_exp}.
  }
  \hfill $\blacksquare$
  \end{remark}

\begin{figure*}[ht!]
  \centering
  \includegraphics[width=0.9\linewidth]{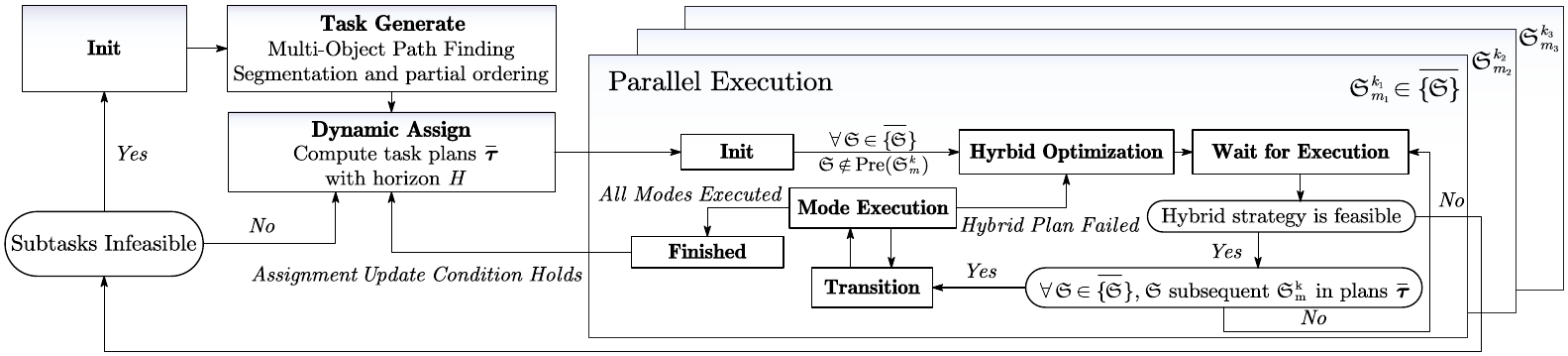}
  \vspace{-0mm}
  \caption{{
      The proposed scheme of online execution
    and adaptation to uncertainties, delays and failures,
    as detailed in Sec.~\ref{subsec:online}.
    Note that the pushing subtasks of different objects are executed in parallel.
  Adaptation is triggered first within the subgroup and then the whole group.}}
  \label{fig:online}
  \vspace{-4mm}
\end{figure*}

\subsubsection{Online Adaptation upon Failures}\label{subsubsec:adapt}
As discussed earlier, due to model uncertainties such as
communication delays, actuation noises and slipping,
the collaborative pushing task is inherently uncertain in terms of both object trajectory
and task duration.
In addition, congestion or deadlock during navigation can also lead to failed execution.
Thus, online adaptation to such contingencies is essential for both the task assignment and the hybrid search
of the pushing strategy.
{
More specifically, the execution of the~$\ell$-th segment of the current hybrid
plan~$\vartheta^{k,\star}_m[\ell] = (\mathbf{s}_\ell,\boldsymbol{\xi}_\ell)$
is considered to be \emph{failed} if one of the following conditions holds:
\begin{equation}\label{eq:fail-cond}
  \begin{split}
&\mathrm{dist}\big{(}\mathbf{s}_m(t'),\, \varrho_\ell\big{)} \geq \delta_{\texttt{f}},\;
    \exists t' \in [t,\,t-T_{\texttt{c}}];\\
&\|\mathbf{s}_m(t') - \mathbf{s}_m(t'')\| < r_{\texttt{stuck}},\;
    \forall t', t'' \in [t,\,t-T_{\texttt{c}}];
  \end{split}
\end{equation}
where~$\mathrm{dist}(\cdot)$ measures the minimum distance
from a point to a 2D curve;
the first condition checks whether the object deviates from the target arc~$\varrho_\ell$
by more than~$\delta_{\texttt{f}}>0$ over a recent time window~$T_{\texttt{c}}>0$;
the second detects if the object is stuck,
i.e., its position change remains below a threshold~$r_{\texttt{stuck}}>0$
during the same period.
Note that $\delta_{\texttt{f}}$ bounds the tracking error,
while setting $\delta_{\texttt{f}}$ too small would trigger frequent replanning
and reduce efficiency.
Moreover, if the object is too close to obstacles,
or new obstacles appear, or several robots fail,
the hybrid search algorithm in Alg.~\ref{alg:hybrid}
is re-executed with the current system state and functional robots,
yielding an adapted hybrid plan $\vartheta^{k,\star}_m$
},
and the control policy in~\eqref{eq:control} is re-activated.
Last but not least,
in case these measures can still not resolve the failure,
the high-level task assignment~$\overline{\boldsymbol{\tau}}$ is updated
by receding horizon planning with the current system state and remaining subtasks.
This can be effective when several robots failed and the remaining robots
in the subgroup are not sufficient to push the object.
The above procedures of online execution and adaptation
are summarized in Fig.~\ref{fig:online}.

%% file: contents/general.tex
\subsection{Discussion}\label{subsec:discuss}

\begin{table}[t!]
\color{black}
\centering
\caption{Conditions for Completeness}
\label{tab:assumption}
\vspace{-2mm}
\begin{tabular}{>{\centering\arraybackslash}p{0.12\linewidth} p{0.78\linewidth}}
\toprule
\multicolumn{1}{c}{\textbf{Cond.}} & \multicolumn{1}{c}{\textbf{Description}} \\
\midrule
C1 & The geometric and physical properties of the objects and robots are known in advance. \\
C2 & Inflating each target object by $(1+\epsilon_{\texttt{r}})$ times the maximum robot diameter, for arbitrarily small $\epsilon_{\texttt{r}}$, yields a feasible multi-agent path finding (MAPF) instance. \\
C3 & For every target object $\Omega_m$, there exists a mode-sufficient robot subgroup $\mathcal{N}_m\subseteq\mathcal{N}$ (see Appendix~\ref{app:sufficient}). \\
\bottomrule
\end{tabular}
\vspace{-4mm}
\end{table}

\subsubsection{Computation Complexity}\label{subsec:computation}
{
The task-level complexity is dominated by MAPF, task decomposition, and task assignment.
For $M$ polygonal objects, a complete MAPF algorithm such as CBS~\cite{sharon2015conflict} is exponential, so we adopt a sequential $\text{A}^\star$ scheme with complexity
$\mathcal{O}(\|V_{\texttt{st}}\|\log\|V_{\texttt{st}}\|\cdot M^2)$,
where $\|V_{\texttt{st}}\|$ is the size of the space-time graph.
Task decomposition and assignment cost
$\mathcal{O}(M^2)$ and $\mathcal{O}(M^H N^{\min(M,H)})$, respectively.
For the hybrid search in Alg.~\ref{alg:hybrid}, the cost is mainly determined by mode generation and the search-tree size.
The feasibility check via $\mathrm{J}_{\texttt{MF}}$ in~\eqref{eq:multi-directional-feasibility}
corresponds to a linear program with complexity $\mathcal{O}(N^{3.5})$~\cite{terlaky2013interior}.
Overall, the time complexity of the hybrid search is
$\mathcal{O}\!\left(N^{3.5} W_k^{2\alpha/\alpha_{\min}}\right)$,
where $W_k$ is the maximum number of nodes generated in a single expansion.
}

{Lastly, as summarized in Table~\ref{tab:assumption},
the key assumptions include that: the intrinsic properties of robots and objects are known,
a feasible MAPF solution exists for all objects,
and there is always a mode-sufficient subgroup of robots for each object.
Then, the completeness guarantee of the overall scheme is stated below,
with detailed proofs in Appendix~\ref{app:proof}.}
  \begin{theorem}\label{theo:all}
    {
      Under the conditions in Table~\ref{tab:assumption},
      the proposed hybrid optimization scheme
      yields valid solutions to Problem~\ref{eq:problem}.
    }
  \end{theorem}

\subsubsection{Generalization}\label{subsec:general}
The proposed framework can be generalized in the following notable directions:

(I) \textbf{Movable obstacles}.
Movable obstacles can be treated as additional objects that are pushed in the same way as targets, 
either to unblock infeasible scenarios or to shorten overall task duration.
They are included in the MAPF instance without fixed goal positions, 
so that timed paths implicitly specify where and how they should be moved, and the resulting subtasks include pushing these movable obstacles.
(II) \textbf{Heterogeneous robots}.
When robots have different capabilities, the high-level assignment should account for their efficiency: 
heavier objects are allocated to more powerful robots, while lighter ones are handled by smaller teams.
The interaction mode can be extended to
$\boldsymbol{\xi}\triangleq(\mathbf{c}_1,\mathbf{f}_1,R_1)\cdots(\mathbf{c}_N,\mathbf{f}_N,R_N)$,
so that contact points $\mathbf{c}_n$ and forces $\mathbf{f}_n$ reflect individual limits, 
and the transition cost $\mathrm{J}_{\texttt{sw}}(\cdot)$ in~\eqref{eq:est-cost} is adapted accordingly.
(III) \textbf{6D pushing tasks}.
For 6D object poses in 3D scenes, the framework extends under a quasi-static, zero-gravity setting.
The 2D arc transition becomes a 3D spiral transition, and the hybrid search space is lifted to 6D pose, 
while the overall structure of the pipeline remains unchanged.
(IV) \textbf{Planar assembly}.
When only a global assembly pattern is specified, 
goal poses for individual objects can be generated by geometric solvers or generative methods~\cite{yamada2024generative}.
The resulting object goals are then passed to our pipeline to complete the assembly.

\begin{figure*}[t!]
  \centering
  \includegraphics[width=0.9\linewidth]{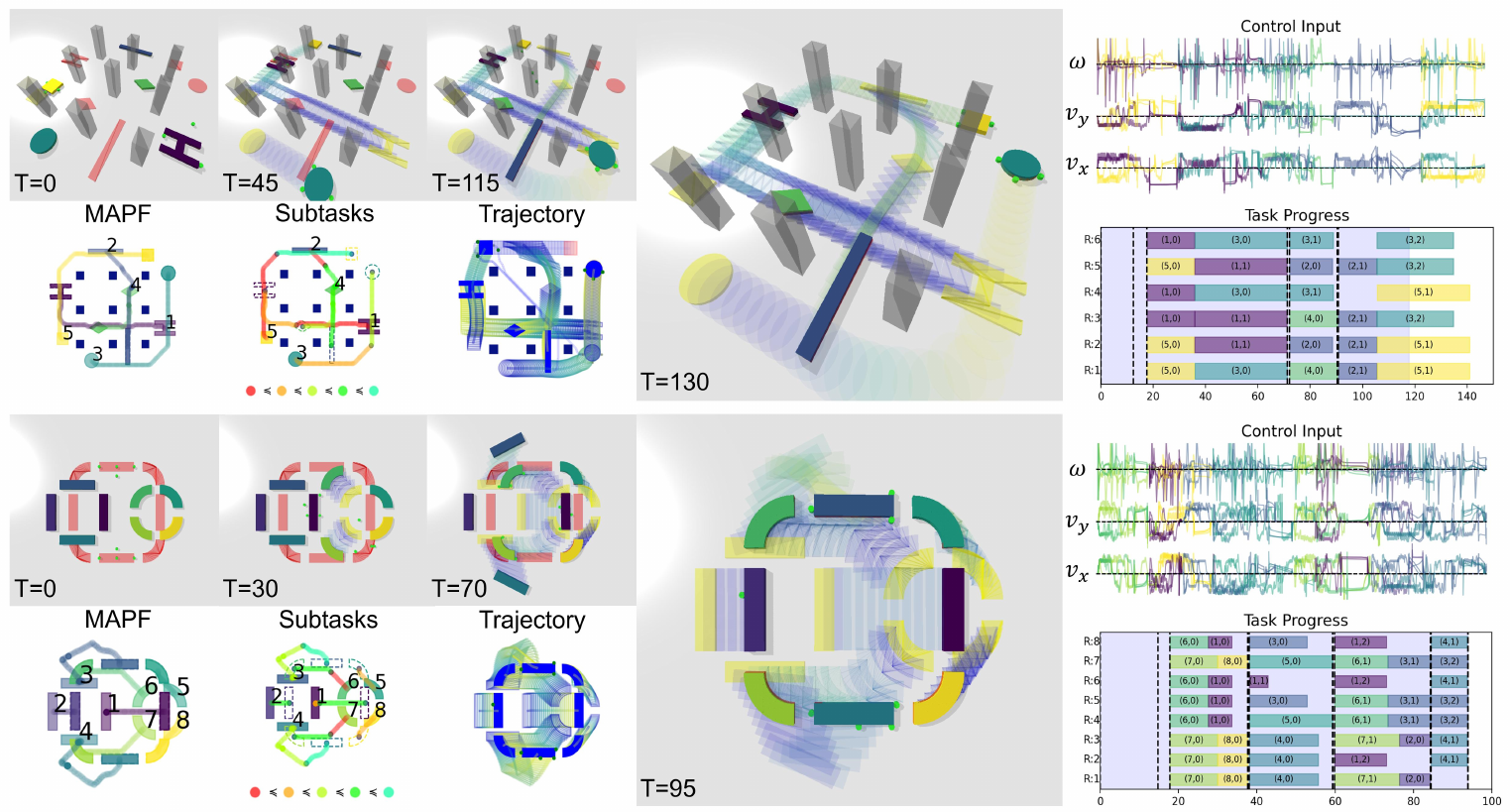}
  \vspace{-0.1in}
  \caption{
  Simulation results of the proposed PushingBot system in two scenarios,
  including the timed path~$\{\mathfrak{S}_m\}$,
  generated subtasks with strict partial order
  (numbered and in different colors),
  hybrid plans and final trajectories~$\mathbf{S}^\star$,
  robot control inputs,
  and the task overall assignment results
  (indexed by the object ID and order of segments).
  \textbf{Top}: $5$ objects (diamond, circular, triangular, square and H-shape), $9$ subtasks,
  and $8$ robots within a forest-like workspace;
  \textbf{Bottom}: $8$ objects ($4$ rectangular
  and~$4$ quarter-circle), $13$ subtasks and $8$ robots.}
  \label{fig:2d_scenarios}
  \vspace{-4mm}
\end{figure*}

\subsubsection{Limitation}\label{subsec:limit}
The proposed approach has several limitations.
(I) As discussed in Appendix~\ref{app:physics}, the quasi-static analysis of collaborative pushing modes holds only for slow motions with small inertia and persistent contact.
For microgravity or floating objects, our method can still operate by actively decelerating, but this can be inefficient and calls for further study.
(II) The mass, shape, and friction coefficients of all objects are assumed known for feasibility evaluation.
Small deviations can be absorbed by the tracking controller in~\eqref{eq:control}, whereas large uncertainties would require active identification and multi-objective planning.
(III) The task decomposition relies on a centralized MAPF solver, and the hybrid search is executed centrally for each subteam.
For large-scale scenarios, decentralized MAPF approaches~\cite{leet2022shard} could reduce complexity, and designing a fully decentralized pushing scheme remains an interesting direction for future work.

%% file: contents/experiment.tex
\section{Numerical Experiments} \label{sec:experiments}
To further validate the proposed method,
extensive numerical simulations and hardware experiments
are presented in this section.
The proposed method is implemented in \texttt{Python3} and tested on a
laptop with an Intel Core i7-1280P CPU.
{
Numerical simulations are run in \texttt{PyBullet}~\cite{coumans2019},
while~\texttt{ROS} is adopted for the hardware experiments.}
The \texttt{GJK} package from~\cite{bergen1999fast} is used for collision checking,
and the \texttt{CVXOPT} package from~\cite{diamond2016cvxpy} for solving linear programs.
Simulation and experiment videos are available
in the supplementary material.
\subsection{Numerical Simulations}\label{subsec:sim}
\subsubsection{{Setup of Simulation Environments}}
\label{subsec:sim-setup}
All simulations are conducted in \texttt{PyBullet}
with a fixed integration time step of~$\Delta t = \frac{1}{240}$\,s.
Unless otherwise specified, the target object has a mass of~$10$kg,
a ground friction coefficient of~$0.8$, and a side friction coefficient of~$0.2$.
Robots are homogeneous cylinders of~$0.3$m diameter and have a maximum pushing force of~$100$N.
A task is considered complete when the object is within~$0.05$m and~$0.15$rad of the goal pose.

\textbf{Control Scheme.}
As discussed in Sec.~\ref{subsubsec:control}, each robot is controlled by a two-level scheme.
The desired velocity to track the arc segments
is given by~$\mathbf{u}_n = (\widehat{\mathbf{v}}_n, \widehat{\omega}_n)$
as in~\eqref{eq:control},
with gains $K_{\texttt{vel}}=5$ and $K_{\texttt{rot}}=1$.
A low-level proportional controller then generates the forces
$\mathbf{f}_{n,\texttt{drv}}=400(\widehat{\mathbf{v}}_n-\mathbf{v}_n)$ and
the torques $\chi_{n,\texttt{drv}}=20(\widehat{\omega}_n-\omega_n)$, which are clipped by the
actuation limits and applied through the force/torque API of \texttt{PyBullet}.
The default robot-ground friction is disabled,
so robots move only under the commanded and clipped forces/torques.
Instead of explicitly modeling the contact dynamics of Mecanum wheels~\cite{zeidis2019dynamics},
this formulation serves as an abstraction of the ideal omni-directional platform,
while the actuation limits bound the realizable pushing forces.

\textbf{Physics Interaction.}
\texttt{PyBullet}
resolves contacts via a sequential-impulse scheme that approximates a
Mixed Linear Complementary Problem (MLCP) under the non-penetration
and Coulomb friction constraints.
However, its lateral and spinning friction are modeled independently,
which is inconsistent with the coupled limit-surface friction model
in the proposed mode-generation module,
and may under-represent the coupled nature of real-world contact.
To mitigate this,
the object-ground friction of \texttt{PyBullet} is replaced by
the ellipsoidal limit-surface model in the force-moment space~\cite{lynch1992manipulation},
with more implementation details in Appendix~\ref{app:quasi-static}.
A sensitivity study towards different friction models
is provided in the supplementary material,
including the relative ranking of all methods and the main conclusions remain unchanged.

\textbf{Algorithm configuration.}
The planning horizon for the dynamic task assignment is set to~$4$.
Assignments are re-evaluated once the first two tasks are completed or after~$80$s.
The tracking threshold~$\delta_{\texttt{c}}$ is set to~$0.1$m.
The desired velocity is updated at~$60$Hz
while low-level controllers are executed at~$240$Hz,
synchronized with the simulation step.
More details are provided in the supplementary material.

\begin{table}[t!]
\vspace{-2mm}
\color{black}
\centering
\caption{Summary of Results for Nominal Scenarios}
\label{tab:results}
\vspace{-0.1in}
\begin{tabular}{lccc}
\toprule
\textbf{Metric} & \textbf{Scen. I} & \textbf{Scen. II} & \textbf{Scen. III}\\
\midrule
Objects/Robots & 4/6 & 5/6 & 8/8 \\
MAPF Time (s) & 9.6 & 12.2 & 15.3\\
Subtasks Generated & 9 & 9 & 13 \\
Longest Task Dependency & 5 & 5 & 5 \\
Task Assignment Time ($s$) & 2.0 & 2.1 & 2.5 \\
Hybrid Optimization Time ($s$) & 0.4 & 0.4 & 0.4 \\
Pushing Modes & 12 & 20 & 18 \\
Completion Time ($s$) & 70 & 130 & 95 \\
Total Planning Time ($s$) & 16.5 & 23.5 & 28.3 \\
\bottomrule
\end{tabular}
\vspace{-2mm}
\end{table}
\subsubsection{Nominal Scenarios}\label{subsec:results}
As shown in Fig.~\ref{fig:overall} and~\ref{fig:2d_scenarios},
three distinctive scenarios are tested:
(I)~$4$ objects (T-shape, cylinder, semi-circular ring, desk-like) are pushed by $6$ robots;
(II)~$5$ objects (circular, triangular, square, H-shape, and diamond)
are pushed through passages between dense prismatic obstacles by $6$ robots to their goals;
(III)~$8$ objects (4 rectangular bars and 4 quarter-circle segments)
are rearranged from a square and circular ring to a rounded square by $8$ robots.
{The proposed framework successfully
completes all pushing tasks in all three scenarios,
with the planning time of each component
summarized in Table~\ref{tab:results}.
The solution time for MAPF increases with the number of objects,
while the longest dependency chain remains constant at $5$ across
three scenarios as primarily determined by path overlap.
The time for the task assignment
remains consistent, due to the receding-horizon scheme.
Moreover, the planning time for the
hybrid optimization per subtask across all objects
takes around~$0.4s$, which is decoupled from overall
task complexity due to the neural accelerated scheme.
The task completion time is collectively influenced
by total path length, the task dependency,
and the object intrinsics.
}

\subsubsection{Generalization}\label{subsec:generalization_exp}
As discussed in Sec.~\ref{subsec:general}
several notable generalizations of the proposed method
are evaluated.

\begin{figure*}[ht!]
  \centering
  \includegraphics[width=0.9\linewidth]{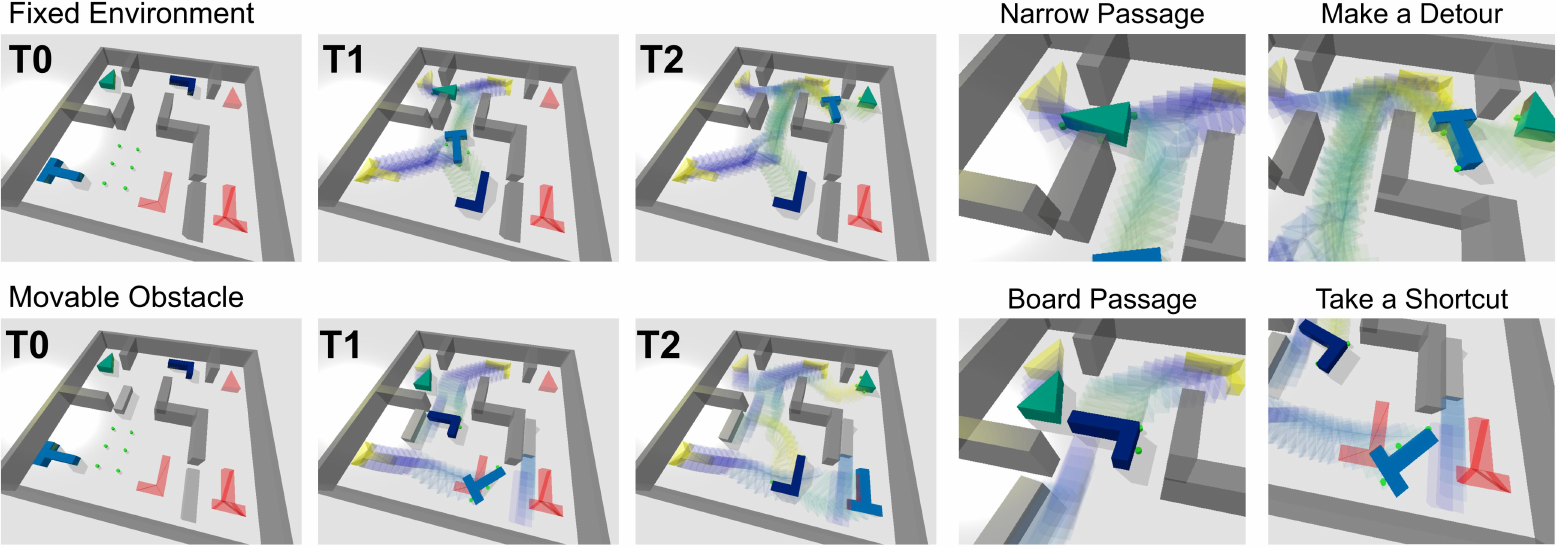}
  \vspace{-0.1in}
  \caption{{
      Generalization of the proposed method to scenarios
      with movable obstacles:
      all obstacles are treated as fixed (\textbf{Top});
      movable obstacles, shown in lighter gray, are pushed to
      create wider passage for both triangular and L-shape objects
      and provide a shortcut for the T-shape object.
      (\textbf{bottom}).}
  }
  \label{fig:movable}
  \vspace{-2mm}
\end{figure*}

\begin{figure*}[t!]
  \centering
  \includegraphics[width=0.9\linewidth]{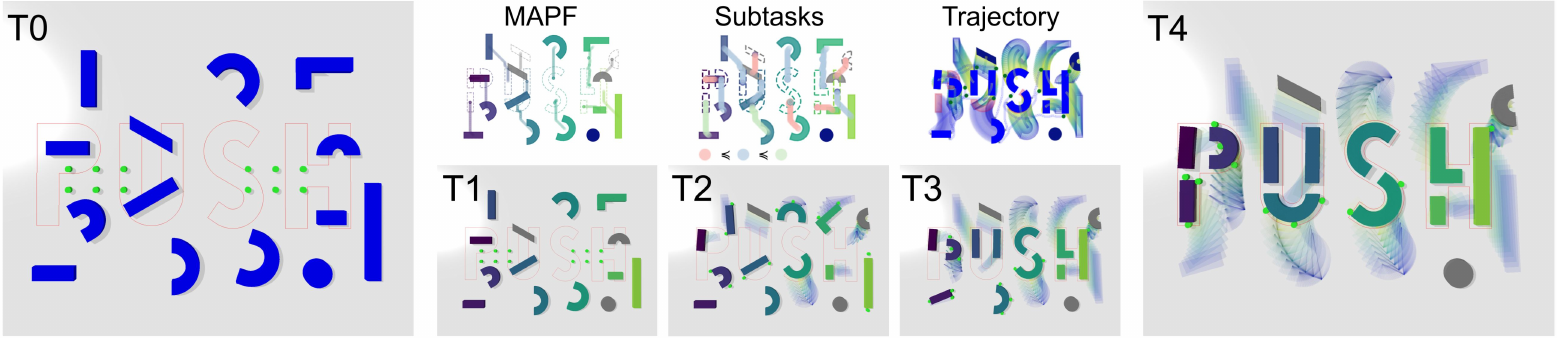}
  \vspace{-0.1in}
  \caption{Generalization of the proposed method
    to planar {assembly} with~$14$ objects
    and~$12$ robots.
    Note that~$11$ objects are selected and
    pushed to suitable target positions,
    while the other objects (in grey) are treated
    as movable obstacles.}
  \label{fig:push}
  \vspace{-4mm}
\end{figure*}

(I) \textbf{Movable obstacles}.
As shown in Fig.~\ref{fig:movable},
$3$ target objects are pushed by~$6$ robots in a $20m \times 20m$ workspace
with fixed (dark grey) and movable (light grey) obstacles.
If movable obstacles are treated as fixed (in the top row of the figure),
the objects must take inefficient detours and narrow passages,
yielding longer MAPF planning times, more complex trajectories,
more frequent mode switching, and increased task duration.
In contrast, treating the movable obstacles as pushable objects without predefined goals
allows our framework to push these obstacles into the nearest freespaces,
creating wider passages and shortcuts for the target objects.
This generalization reduces the task duration from~$83.2s$ to $63.3s$ by~$23\%$,
while retaining safety and liveness.

\begin{figure}[t!]
  \vspace{-1mm}
  \centering
  \includegraphics[width=0.9\linewidth]{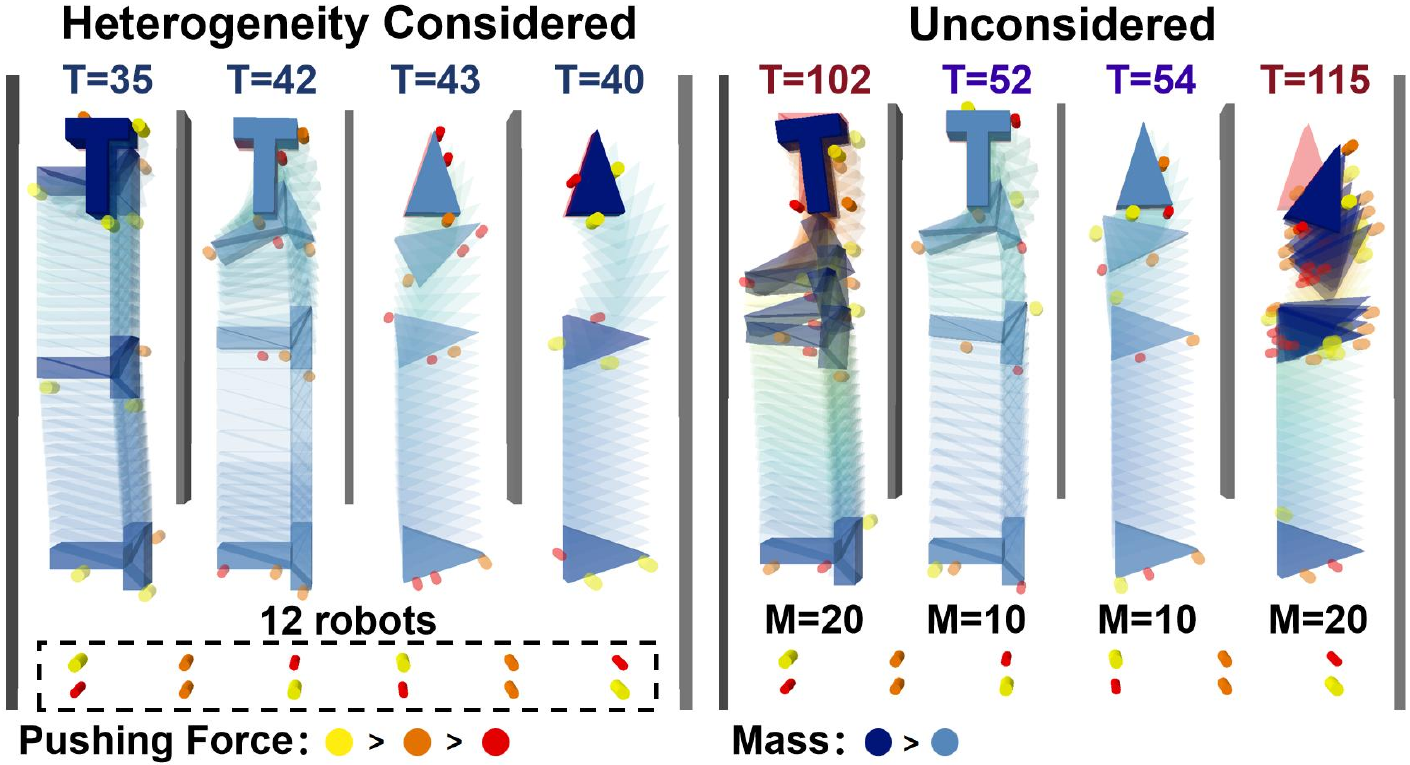}
  \vspace{-0.1in}
  \caption{
    Generalization of the proposed method to~$12$ heterogeneous
    robots with varying maximum pushing forces,
    and~$2$ objects with different masses:
    heterogeneity is considered (\textbf{Left}) and neglected (\textbf{Right})
    in the hybrid optimization,
    resulting in significantly different execution times.
  }
  \label{fig:heterogeneous}
  \vspace{-2mm}
\end{figure}

\begin{figure}[t!]
  \centering
  \includegraphics[width=0.9\linewidth]{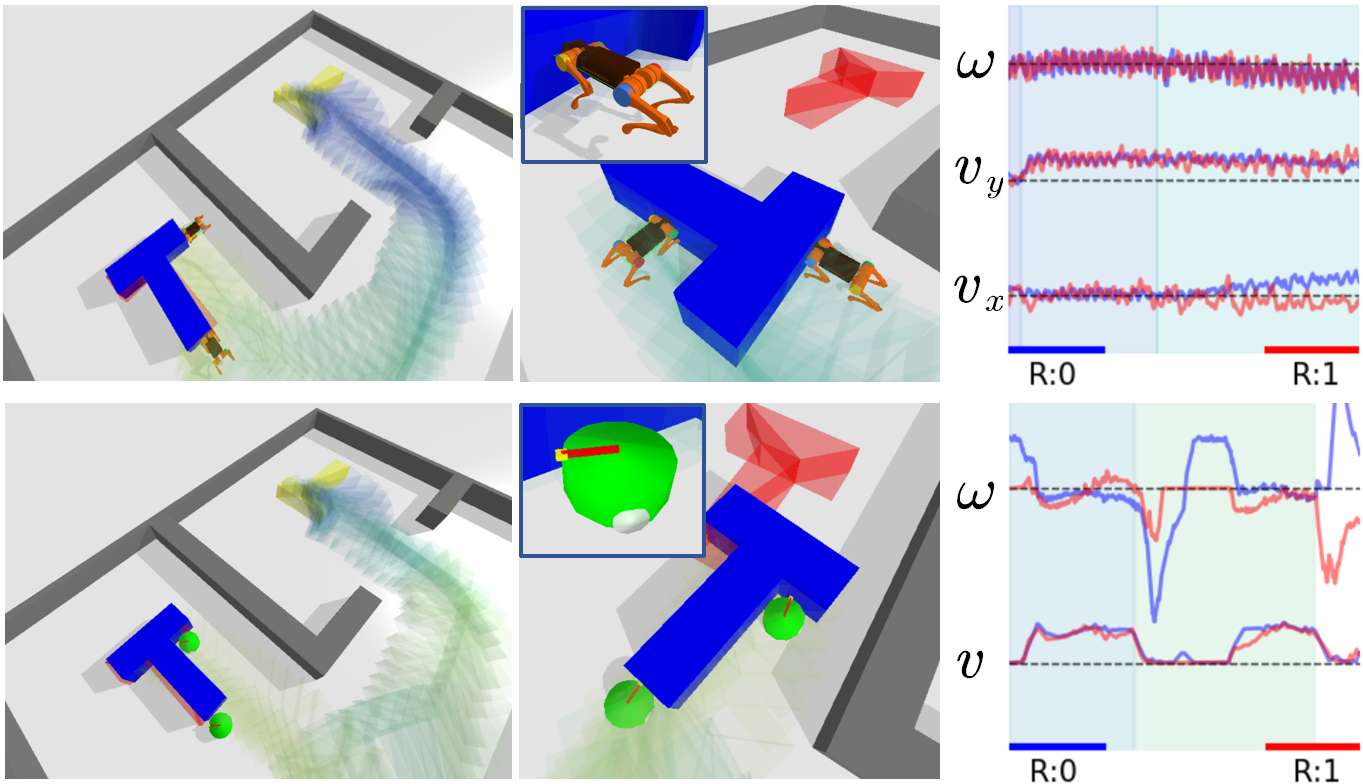}
  \vspace{-0.1in}
  \caption{
    {
    Generalization of the proposed method to
    non-holonomic robots such as
    quadruped and differential-drive robots.
    Snapshots along with velocity control inputs are also shown.
    }
  }
  \label{fig:robot_extention}
  \vspace{-6mm}
\end{figure}

(II) \textbf{Heterogeneous robots}.
A scenario of pushing race is designed to demonstrate the capability of
the proposed method to handle heterogeneous robots and objects.
As shown in Fig.~\ref{fig:heterogeneous},
the setup involves two T-shape objects
and two triangular objects.
For each shape, one object has a mass of~$5kg$ and the other has a mass of~$20kg$.
These objects are placed
in a row at one end of a~$20m \times 20m$ workspace,
with the target position at the opposite end.
In total, $6$ heterogeneous robots are deployed, i.e.,
the maximum pushing forces~$f_{n,\max}$
in~\eqref{eq:force-limit} are $100N$, $30N$ and $10N$
and each force level is associated with two robots.
The assignment and planning strategy
as described in Sec.~\ref{subsec:general} is adopted,
by which all tasks are completed with different duration
but all below~$40s$.
Compared to the nominal method which neglects these heterogeneity,
the assignments and pushing modes are intuitive, i.e.,
(i) heavier objects are always pushed by more and stronger robots,
while lighter objects are handled by the remaining robots.
(ii) robots with larger forces are always assigned to
the contact point that requires more force.
Although the task can still be completed via the nominal method,
the task duration is much longer (by minimum~$23\%$ and maximum~$188\%$).
Last but not least,
as shown in Fig.~\ref{fig:robot_extention},
quadruped and differential-drive robots are adopted
for the collaborative pushing task.
We adopt a soft ``turn-and-forward'' scheme that smoothly scales linear speed by the
heading error, which suffices for our task settings.
For higher-precision pushing with nonholonomic platforms (e.g., centimeter-level path tracking),
a more foundational treatment such as Jourdain's principle~\cite{ebel2024cooperative} may be required.

(III) \textbf{Planar Assembly}.
As shown in Fig.~\ref{fig:push}, a planar assembly task of assembling
the word ``PUSH'' via~$14$ objects and~$12$ robots is demonstrated.
The desired layout is highlighted in red at the center of a~$26m \times 20m$ area,
with objects initially randomly placed.
Heuristic programming method in~\cite{deutsch1972heuristic}
is adopted to select objects and optimize their goal positions.
For instance, two identical semicircular ring-like objects are chosen
to form the letters ``S'',
while L-shaped objects can be part of the letter ``H'', and so on.
Objects that are not required in the assembly,
such as parallelograms or cylinders, are treated as movable obstacles.
The selected objects and their timed paths are shown
in Fig.~\ref{fig:push}.
Once this procedure is completed, the proposed scheme
is applied to push the selected objects to their goal positions.
The whole assembly is completed at~$t=108s$,
during which~$13$ subtasks and $19$ pushing modes and $6$ mode switches
are performed.

\begin{figure*}[ht!]
  \centering
  \includegraphics[width=0.9\linewidth]{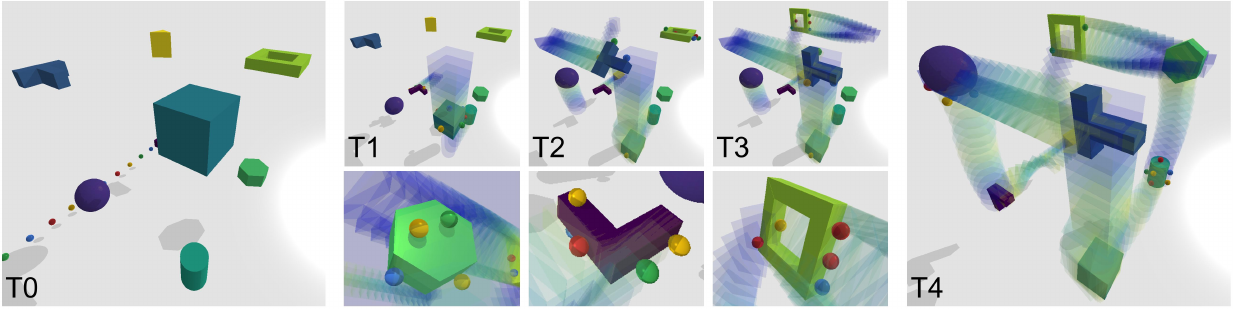}
  \vspace{-0.1in}
  \caption{
    Generalization of the proposed method
    to 6D pushing tasks, where
    the positions of~$8$ objects are swapped
    via~$12$ robots.
    Note that the robots switch subtasks frequently
  due to the cluttered workspace
  and the large number of partial order dependencies between the subtasks.
  }
  \label{fig:6D}
  \vspace{-2mm}
\end{figure*}

(IV) \textbf{6D Pushing}.
As shown in Fig.~\ref{fig:6D},
the generalization to pushing tasks in complex 3D workspace is demonstrated.
The scenario consists of~$8$ objects and~$12$ robots
within a workspace of~$10m\times 10m\times 10m$.
Initially, all objects are placed at one vertex of a cube
and should be pushed to another vertex, i.e., to swap their positions.
The resistance coefficient of all objects are set to~$\mathbf{K}=100$.
Thus, the MAPF algorithm in $3$D is first employed to generate~$8$
timed paths $\{\mathfrak{S}_m\}$ within~$2.3s$,
which are then decomposed into~$13$ subtasks.
All tasks complete at~$t=165s$ with a total planning time of~$15.2s$
and $21$ pushing modes.
The difficulty of stably pushing an object in 3D space
depends on the diversity of its surface normal vectors.
For instance,
the triangular prism has only~$5$ faces,
while the sphere has arbitrary normal vectors.
Consequently,
the proposed method assigns~$6$ robots to the triangular prism,
but only~$3$ robots to the sphere.
\begin{figure*}[t!]
  \centering
  \includegraphics[width=0.9\linewidth]{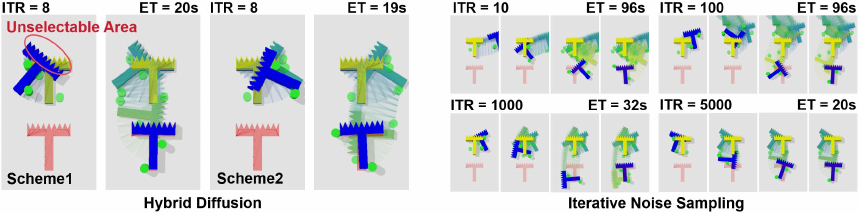}
  \vspace{-2mm}
  \caption{
  {
    Comparison of the diffusion-based method (\textbf{Left}) and iterative sampling (\textbf{Right})
    for generating keyframes and pushing modes.
    The diffusion method generates multi-modal policies in 8 iterations with an execution time of $19$s,
    while iterative sampling requires $5000$ iterations for similar plans.
  }
  }
  \label{fig:noise_sampling}
  \vspace{-2mm}
\end{figure*}
\begin{figure*}[t!]
  \centering
  \includegraphics[width=0.9\linewidth]{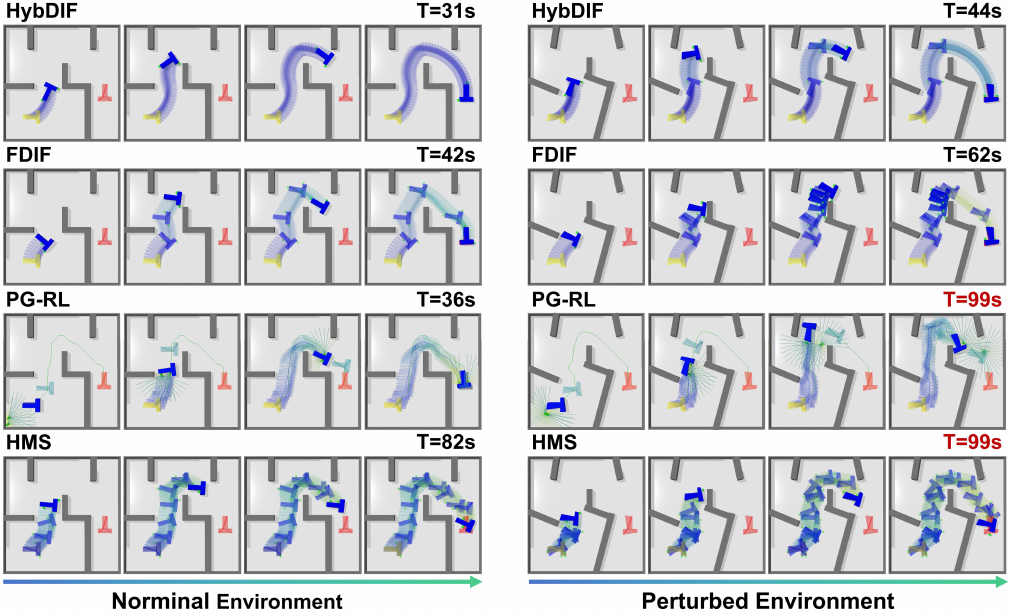}
  \vspace{-2mm}
  \caption{Comparison of the proposed method with~$5$ baselines in the single-object
    pushing task,
    in a nominal environment that are similar to training datasets (\textbf{Left})
    and a perturbed environment where obstacles are tilted
    and objects are deformed (\textbf{Right}).}
  \label{fig:single-comparison}
  \vspace{-4mm}
\end{figure*}
\begin{table}[tp]
\vspace{-3mm}
\color{black}
\centering
\caption{Comparison between diffusion and iterative sampling}
\vspace{-0.1in}
\resizebox{0.9\linewidth}{!}{
  \begin{threeparttable}
  \begin{tabular}{lccc}
  \toprule
  \textbf{Method} & \textbf{ITR}\tnote{1} & \textbf{PT (s)}\tnote{2} & \textbf{ET (s)}\tnote{3} \\
  \midrule
  \textbf{Diffusion (DDPM)} & 100   & 1.60  & 20.4 \\
  \textbf{Diffusion (DDIM)} & 8     & 0.15  & 20.6 \\
  \midrule
  \multirow{4}{*}{\textbf{Samp. (w/ mode library)}}
    & 10    & 0.05 & 46.8 \\
    & 100   & 0.22 & 32.1 \\
    & 1000  & 2.45 & 24.5 \\
    & 5000  & 12.4 & 20.2 \\
  \midrule
  \textbf{Samp. (w/o mode library)} & 5000 & 453 & 21.6 \\
  \bottomrule
  \end{tabular}
  \begin{tablenotes}
  \item[1] Number of iterations or inference steps.
  \item[2] Planning time per trial.
  \item[3] Execution time per trial.
  \end{tablenotes}
  \end{threeparttable}
}
\vspace{-6mm}
\label{tab:performance_comparison}
\end{table}

\subsubsection{Comparison of Diffusion and Iterative Sampling}
We compare the proposed diffusion-based method with the iterative sampling method
in Sec.~\ref{subsubsec:hybrid-search} in terms of sample efficiency and execution quality.
As shown in Fig.~\ref{fig:noise_sampling} and summarized in Table~\ref{tab:performance_comparison},
the diffusion model, especially the DDIM variant,
outperforms iterative sampling in both sample efficiency and execution time.
With only 8 iterations and $0.15$s planning time, DDIM achieves a comparable execution time of $20.6$s
to the iterative sampling method after 5000 iterations,
demonstrating a $35$-fold reduction in iterations.
Iterative sampling with the {mode library}
as introduced in Sec.~\ref{subsubsec:diffusion}
shows diminishing returns beyond 1000 iterations,
with execution time plateauing at around 20s despite a $408$\% increase in planning time.
Without the diffusion policy or the mode library,
the computational cost becomes prohibitive,
requiring~$453$s for planning with only marginal improvement in execution time.
This highlights that diffusion models distill knowledge from high-iteration sampling
into efficient generative models, offering a computationally superior alternative.
Furthermore, the diffusion-generated plans exhibit
execution feasibility comparable to those from iterative sampling,
confirming their accuracy despite a generation time of just $0.15$s.

\subsubsection{Comparison of Single-object Pushing}\label{subsec:single-compare}
We compare the performance of the proposed hybrid optimization scheme (\textbf{HybDIF})
against six baselines in single-object pushing tasks:
(I) \textbf{KGHS}, the keyframe-guided hybrid search without neural acceleration;
(II) \textbf{FDIF}, a diffusion-based algorithm that generates a complete hybrid plan,
which incorporates image observations for cluttered environments;
(III) \textbf{HMS}, which uses heuristic sampling for mode generation
without further evaluation and optimization;
(IV) \textbf{CMTC}, adapted from~\cite{ebel2024cooperative} without {co-optimizing} the path and modes;
(V) \textbf{MARL}, a multi-agent reinforcement learning baseline based on the PPO algorithm~\cite{yu2022surprising};
(VI) \textbf{PG-RL}, an extension of MARL with a guiding path planning module,
from which a local target position is selected for guiding RL agents.

\begin{figure*}[t]
  \centering
  \includegraphics[width=0.9\linewidth]{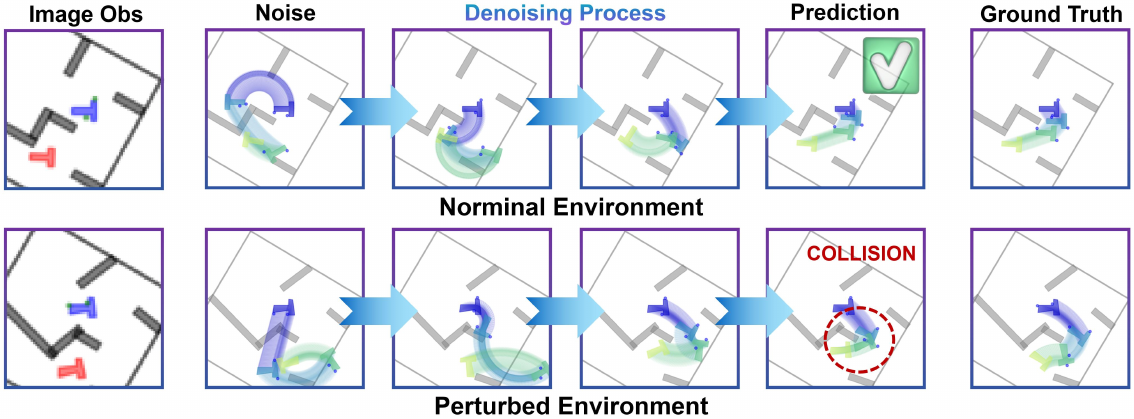}
  \vspace{-2mm}
  \caption{
  The denoising process of the baseline method {FDIF}
  in both the nominal scenario (\textbf{Top})
  and perturbed scenario (\textbf{Bottom}),
  where the predicted keyframes appear in collision within the perturbed scenario,
  which would lead to a higher collision count and frequent replanning
  in online execution of {FDIF}.
  }
  \label{fig:baseline_FHDif}
  \vspace{-2mm}
\end{figure*}
\begin{figure*}[t]
  \centering
  \includegraphics[width=0.9\linewidth]{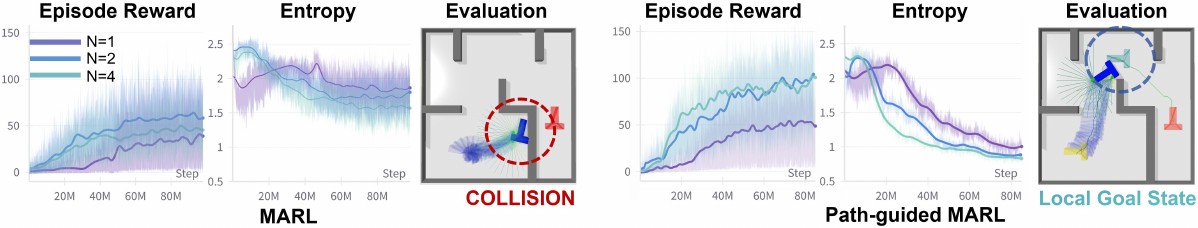}
  \vspace{-0.1in}
  \caption{
    The training and testing results of MARL-based baselines:
    the vanilla MARL~\cite{yu2022surprising} (\textbf{Left})
    and the path-guided PG-RL (\textbf{Right}).
  \label{fig:rl-training}
  }
  \vspace{-3mm}
\end{figure*}
\begin{table*}[t]
  \begin{center}
  \caption{
  {Comparison of single-object tasks with $N=1, 2, 4$ robots
  for nominal and perturbed environments (averaged over~$30$ tests)}
}
 \label{tab:comparison}
  \vspace{-1mm}
  \resizebox{0.9\linewidth}{!}{
  \begin{tabular}{c|c|ccc|ccc|ccc|ccc|ccc} 
    \toprule[1pt]
    \multirow{2}{*}{\textbf{Env}}
    & \multirow{2}{*}{\textbf{Method}}
    & \multicolumn{3}{c|}{\textbf{Success Rate}}
    & \multicolumn{3}{c|}{\textbf{Execution Time}}
    & \multicolumn{3}{c|}{\textbf{Planning Time}}
    & \multicolumn{3}{c|}{\textbf{Control Cost}}
    & \multicolumn{3}{c}{\textbf{Collision Count}} \\

    \cmidrule(lr){3-5} \cmidrule(lr){6-8} \cmidrule(lr){9-11} \cmidrule(lr){12-14} \cmidrule(lr){15-17}
     &  & N=1 & N=2 & N=4 & N=1 & N=2 & N=4 & N=1 & N=2 & N=4 & N=1 & N=2 & N=4 & N=1 & N=2 & N=4 \\

    \midrule
    \multirow{6}{*}{\textbf{Nominal}}
    & \textbf{HybDIF}(ours) &\textbf{1.00}  &\textbf{1.00}  &\textbf{1.00}&\textbf{35.8}  &\textbf{21.5}  &\textbf{22.4}  &\textbf{1.87}  &\textbf{0.88}  &1.02           &\textbf{25.2}  &17.7           &\textbf{13.1}  &\textbf{0.12}  &\textbf{0.03}  &\textbf{0.03} \\
    & \textbf{KGHS}   &0.97           &\textbf{1.00}    &\textbf{1.00}    &38.2           &28.3           &23.2           &4.76           &3.67           &4.25           &26.1           &17.3           &14.2           &0.31           &\textbf{0.03}  &\textbf{0.03} \\
    & \textbf{FDIF}   &0.87           &0.94             &0.97             &46.3           &38.4           &32.5           &3.26           &2.10           &1.25           &31.4           &24.5           &20.7           &0.65           &0.31           &0.25 \\
    & \cb\textbf{CMTC}&\cb0.72 &\cb0.84 &\cb0.97
                      &\cb95.3 &\cb68.7 &\cb50.1
                      &\cb10.5 &\cb4.20 &\cb3.10
                      &\cb80.4 &\cb32.6 &\cb24.9
                      &\cb0.30 &\cb0.04 &\cb0.05 \\
    & \cb\textbf{HMS} &\cb0.93 &\cb\textbf{1.00} &\cb\textbf{1.00}
                      &\cb70.2 &\cb40.8          &\cb34.0
                      &\cb8.52 &\cb2.40          &\cb2.00
                      &\cb76.3 &\cb28.5          &\cb22.4
                      &\cb0.80 &\cb0.20          &\cb0.07 \\
    & \textbf{PGRL}   &0.97           &\textbf{1.00}    &\textbf{1.00}    &47.9           &25.6           &28.3           &3.71           &1.06           &\textbf{0.98}  &27.3           &\textbf{16.8}  &16.2           &0.52           &0.30           &0.27 \\
    & \textbf{MARL}   &0.53           &0.64             &0.55             &67.2           &52.2           &59.4           &2.04           &1.73           &2.12           &40.2           &24.1           &28.2           &0.72           &1.43           &1.17 \\
    \midrule
    \multirow{6}{*}{\textbf{Perturbed}}
    & \textbf{HybDIF}(ours) &\textbf{1.00}  &\textbf{1.00}  &\textbf{1.00}&\textbf{36.4}  &\textbf{28.1}  &\textbf{22.3}  &\textbf{2.04}  &\textbf{0.98}  &\textbf{1.15}  &\textbf{26.3}  &\textbf{17.6}  &\textbf{12.8}  &\textbf{0.22}  &0.06           &\textbf{0.03} \\
    & \textbf{KGHS}   &0.97           &\textbf{1.00}    &\textbf{1.00}    &36.7           &29.5           &23.7           &4.82           &3.92           &4.18           &28.1           &18.3           &15.3           &0.32           &\textbf{0.03}  &\textbf{0.03} \\
    & \textbf{FDIF}   &0.72           &0.82             &0.86             &52.1           &42.3           &37.9           &5.90           &4.92           &2.96           &36.5           &28.9           &25.1           &1.21           &0.60           &0.42 \\
    & \cb\textbf{CMTC}&\cb0.65 &\cb0.76 &\cb0.93
                      &\cb102.4&\cb73.2 &\cb55.8
                      &\cb11.7 &\cb4.60 &\cb3.30
                      &\cb85.7 &\cb34.8 &\cb25.7
                      &\cb0.32 &\cb0.10 &\cb0.08 \\
    & \cb\textbf{HMS} &\cb0.85  &\cb\textbf{1.00} &\cb\textbf{1.00}
                      &\cb82.0  &\cb48.5          &\cb38.1
                      &\cb9.53  &\cb3.05          &\cb2.34
                      &\cb87.9  &\cb32.6          &\cb23.0
                      &\cb0.85  &\cb0.26          &\cb0.08 \\
    & \textbf{PGRL}   &0.64           &0.75             &0.82             &64.5           &56.4           &44.3           &2.06           &2.12           &1.94           &38.4           &27.6           &21.6           &1.50           &1.23           &0.87 \\
    & \textbf{MARL}   &0.47           &0.57             &0.52             &65.1           &62.7           &58.1           &1.98           &1.62           &1.19           &39.5           &24.1           &22.1           &1.16           &1.72           &1.34 \\
    \bottomrule[1pt]
  \end{tabular}
  }
  \end{center}
  \vspace{-6mm}
\end{table*}

As shown in Fig.~\ref{fig:single-comparison}, we evaluate two scenarios:
a nominal one with a T-shaped object and obstacles,
and a perturbed one where the object is deformed and obstacles are rotated.
FDIF, RL, and PGRL sample data only in the nominal scenario,
while the perturbed environment is unseen.
In contrast, HybDIF collects data exclusively in free space,
independent of obstacle layouts, and \emph{does not} include the deformed object.
For both scenarios, initial and goal positions are random,
and each algorithm is tested 30 times.
Table~\ref{tab:comparison} reports five metrics
(success rate, execution time, planning time, control cost, collision count)
for $N=1,2,4$ robots.

\begin{figure*}[t!]
  \centering
  \includegraphics[width=0.9\linewidth]{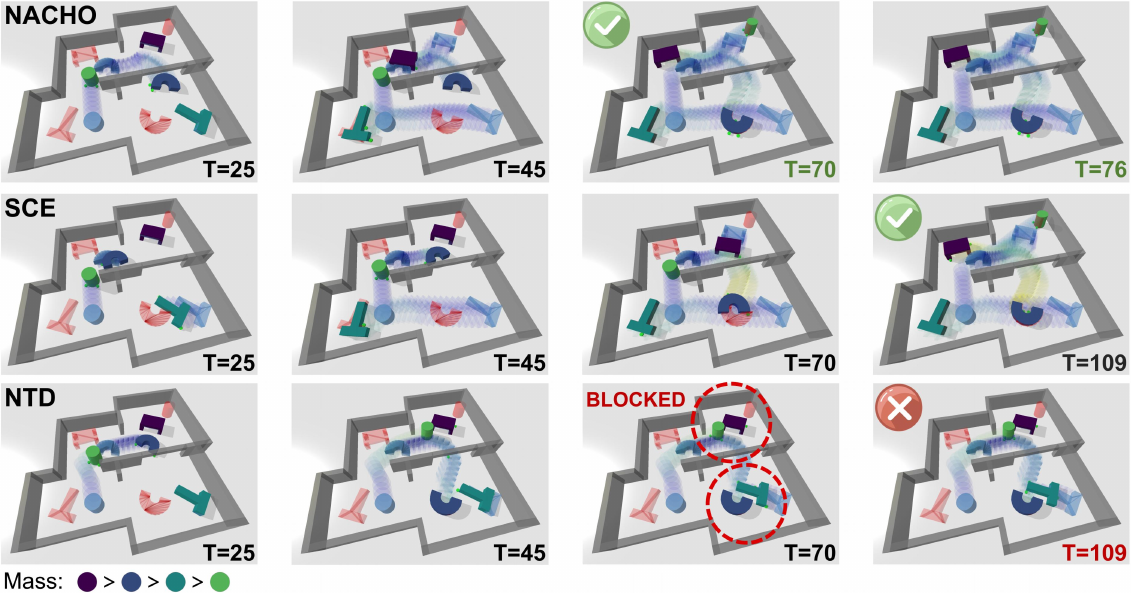}
  \vspace{-0.1in}
  \caption{
    Comparison of different baseline methods for the multi-object pushing tasks,
    where~$4$ objects have different shapes and masses.
  }
  \label{fig:mt_compare}
  \vspace{-4mm}
\end{figure*}

In the nominal scenario,
\textbf{HybDIF} achieves the highest success rate ($100\%$)
with the shortest execution and planning times across all robot numbers.
It also maintains the lowest control cost and collision count,
demonstrating superior overall performance.
{KGHS} also achieves $100\%$ success for~$N=2, 4$.
However, its execution and planning times, along with control costs,
are significantly higher than {HybDIF}.
In contrast, FDIF has a lower success rate
and higher planning time and control costs than {HybDIF} and {KGHS}.
{
{CMTC} achieves a success rate of $96\%$ for~$N=4$,
but for $N=1, 2$, it has a much lower success rate ($72\%$ and $84\%$),
as the smaller matching degree between the path and the contact modes.
}
Although {HMS} achieves a $100\%$ success rate for~$N=2, 4$
it has the longest execution time and highest control cost.
{PGRL} performs competitively with a success rate around~$96\%$,
but its control cost and collision count exceed {HybDIF}.
Lastly, {MARL} has the lowest success rate and highest collision count,
indicating poor performance for long-distance pushing tasks.
As analyzed in Fig.~\ref{fig:rl-training},
PGRL reduces the difficulty of learning long-distance pushing strategies
by selecting local goals.

More importantly, in the perturbed scenario,
{HybDIF} maintains the best performance with a~$100\%$ success
rate, the shortest execution time and the fastest planning time across all robot numbers,
validating its robustness to perturbations in obstacle layout and object shape.
In contrast, {FDIF} experiences a significant drop in success rate.
As shown in Fig.~\ref{fig:baseline_FHDif},
the diffusion model generates valid plans effectively in the nominal scenario.
However, the generated keyframes are often \emph{in collision} within the perturbed scenario
due to insufficient training data that fails to cover
a diverse range of obstacle distributions and object shapes.
yielding a much higher collision count by~$720\%$.
Similar degradation can be found in {MARL} and {PGRL},
with a reduced success rate (by~$10\%$ and $26\%$)
and increased collision count (by~$27\%$ and $233\%$).
Though {PGRL} still maintains an advantage compared to {MARL},
it experiences a more significant performance drop,
indicating its vulnerability to the environment perturbations.

\begin{figure}[t!]
  \centering
  \includegraphics[width=0.95\linewidth]{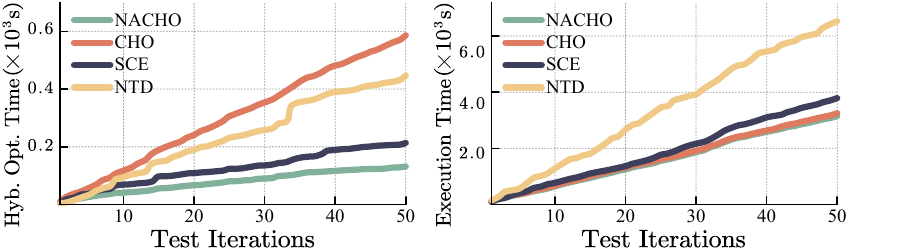}
  \vspace{-0.1in}
  \caption{
    Accumulated planning time (\textbf{Left})
    and accumulated execution time (\textbf{Right})
    over~$50$ multi-object pushing tasks.
  }
  \label{fig:mt_cumu_times}
  \vspace{-4mm}
\end{figure}

\subsubsection{Comparison of Multi-object Pushing}\label{subsec:multi-compare}
To further evaluate the performance of the proposed neural accelerated
combinatorial-hybrid optimization method (\textbf{NACHO}),
the following three baselines are considered:
(I) \textbf{CHO},
which uses the same model-based task assignment and hybrid optimization modules as NACHO,
but without diffusion-based neural acceleration;
(II) \textbf{SCE},
which disregards the heterogeneity of robots and objects and instead assumes uniform capabilities,
e.g., identical object weights and robot force limits in both task assignment and mode generation;
(III) \textbf{NTD},
which retains the task-allocation algorithm of~\cite{choudhury2022dynamic}
but omits the MAPF-based sub-task decomposition and partial-order construction,
so the robots push the object directly along the raw MAPF path.

Fig.~\ref{fig:mt_compare} shows snapshots of one execution in the nominal scenario with varying object masses and shapes.
NACHO completes the task first at $t=70$s,
while CHO is slower without neural acceleration.
Due to the simplified module of cost estimation, SCE is much slower and often exhibits inefficient behaviors,
e.g., heavier objects such as the table and the semi-circle are assigned to only two robots.
Without the ordering of subtasks,
NTD fails to complete the task as the objects often block each other during execution.

Moreover, a quantitative comparison across 50 tasks in different environments
is provided in Table\ref{tab:mt_compare} and Fig.~\ref{fig:mt_cumu_times}.
MAPF and task decomposition take about $9.1$s across all methods,
while task assignment and hybrid optimization take $1.82$s and $2.64$s for NACHO, respectively.
NACHO achieves the shortest planning ($14.7$s) and execution time ($78.7$s),
outperforming other methods by $37.8\%$.
In contrast, SCE is the slowest with an execution time of $94.7$s,
highlighting the importance of considering robot and object heterogeneity.
NTD has the lowest success rate ($44\%$) and high execution time ($86.2$s),
proving that simply following the MAPF path  is infeasible.

\subsubsection{Geometric Pushing Puzzle for Ants}\label{subsec:ant-compare}
\begin{figure}[t!]
  \centering
  \includegraphics[width=0.9\linewidth]{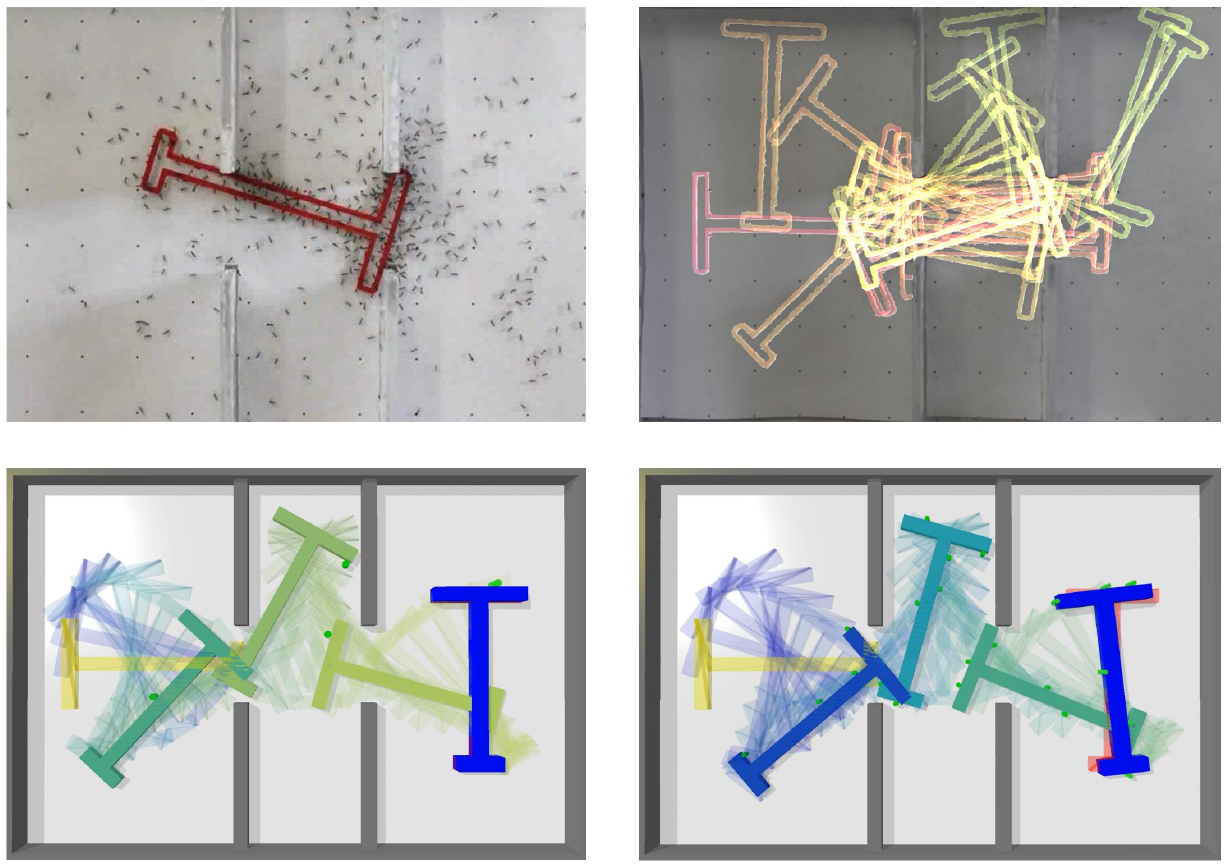}
  \vspace{-0.1in}
  \caption{Hard geometric pushing puzzle proposed in~\cite{dreyer2025comparing}
    from the left side to the right side within the confined space.
    \textbf{Top}: the final object trajectory as pushed by hundreds of ants,
    abstracted from the recorded video in~\cite{dreyer2025comparing};
    \textbf{Down}: the object trajectory via the proposed method
    with~$1$ robot (within~$103$s and~$23$ modes) and~$6$ robots
    (within~$52$s and~$9$ modes).}
  \label{fig:ants}
  \vspace{-4mm}
\end{figure}

A recent study in~\cite{dreyer2025comparing} poses a hard geometric puzzle
for hundreds of ants to \emph{lift, push or pull}
a large dumbbell-shaped object through
two narrow passages in confined space, as shown in Fig.~\ref{fig:ants}.
The ants demonstrate amazing coordination and collaboration to accomplish this task,
despite their limited field of view and local communication.
For (maybe unfair) comparisons,
a similar scenario is replicated, for which~$1$ and~$6$ robots are
deployed to see different strategies.
Note that without the ability to pull,
the robots can still accomplish the task via the proposed scheme.
It takes~$103$s and~$23$ modes for~$1$ robot,
and~$52$s and~$9$ modes for~$6$ robots.
Nonetheless, the observations in~\cite{dreyer2025comparing} have
inspired our future work towards distributed coordination schemes
with only local communication.

\begin{table}[t!]
  \centering
  \caption{Comparison with~$3$ baselines for the multi-object pushing task\label{tab:mt_compare}}
  \vspace{-2mm}
  \resizebox{0.95\linewidth}{!}{
    \begin{threeparttable}
        \begin{tabular}{cccccc}
        \toprule[1pt]
        \textbf{Algorithm}
        & \textbf{SR}\tnote{1}  & \textbf{MDT}\tnote{2}
        & \textbf{TAT}\tnote{3} & \textbf{HOT}\tnote{4}
        & \textbf{ET}\tnote{5}\\
        \midrule
        \textbf{NACHO}(ours)& \textbf{1.00} \   &10.3 \     &1.78 \     &\textbf{2.64} \ &\textbf{78.7} \ \\
        \midrule
        \textbf{CHO}        & \textbf{1.00} \   &10.2 \     &1.78 \     &11.7 \       &81.3 \ \\
        \midrule
        \textbf{SCE}        & \textbf{1.00} \   &10.3 \     &1.83 \     &4.27 \       &94.7 \ \\
        \midrule
        \cb\textbf{NTD}     &  \cb 0.45 \       &\cb 9.05 \ &\cb 1.65 \ &\cb 8.26 \   &\cb 89.7 \ \\
        \bottomrule[1pt]
        \end{tabular}
      \begin{tablenotes}
      \item[1] Success rate.
      \item[2] Time for task decomposition.
      \item[3] Time for task assignment.
      \item[4] Time for hybrid optimization.
      \item[5] Execution time.
      \end{tablenotes}
      \vspace{-2mm}
    \end{threeparttable}
  }
\end{table}

\subsection{Hardware Experiments}\label{subsec:hardware}

\subsubsection{System Description}\label{subsec:exp-description}
\begin{figure*}[t!]
  \centering
  \includegraphics[width=0.95\linewidth]{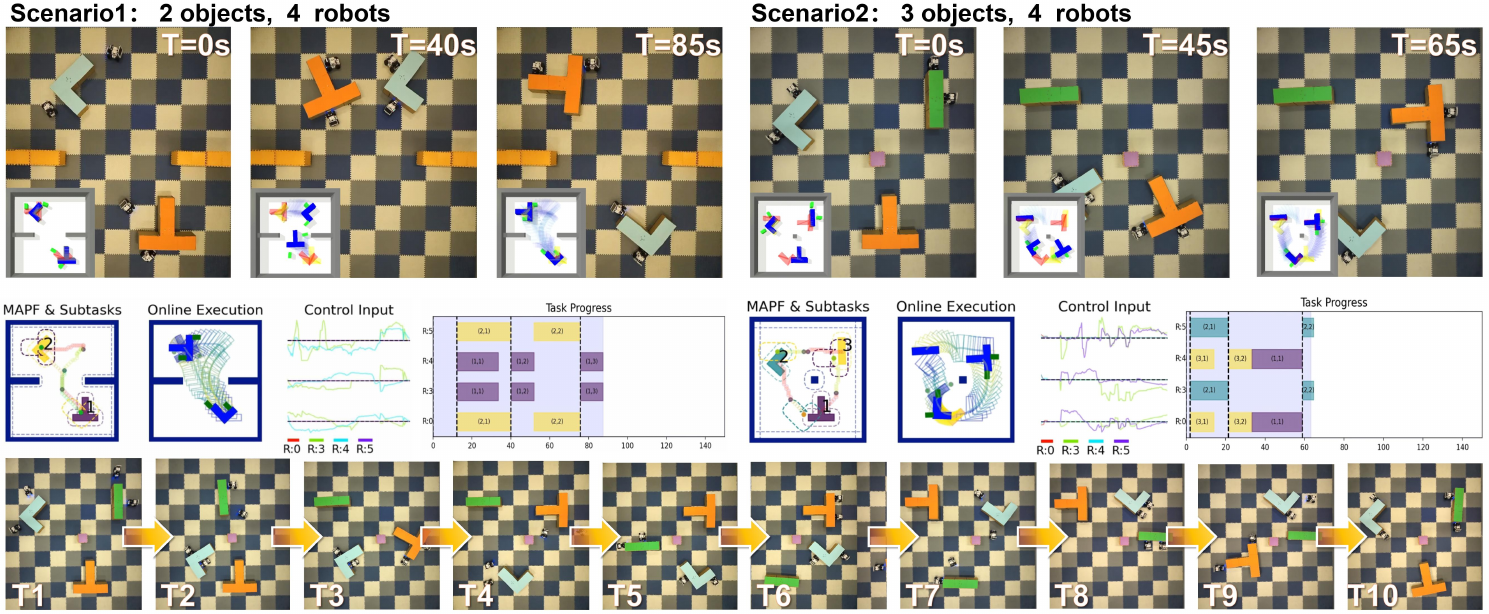}
  \vspace{-2mm}
  \caption{Snapshots of the PushingBot system during hardware experiments,
    including the pushing subtasks, the hybrid plan, control inputs and Gantt graph
    of subtasks.
    \textbf{Left}: four robots are deployed for swapping two objects in Scenario-I;
    \textbf{Right}: four robots are deployed to rotate three objects in Scenario-II.}
  \label{fig:hardware}
  \vspace{-4mm}
\end{figure*}

As shown in Fig.~\ref{fig:hardware},
the experiments are conducted in a~$5m \times 5m$ lab with the \texttt{OptiTrack}
motion capture system for global positioning.
Four identical Mecanum-wheel robots are deployed,
each measuring $0.2\mathrm{m} \times 0.3\mathrm{m}$
with a maximum pushing force of~$10\mathrm{N}$.
Each robot is equipped with a NVIDIA Jetson Nano running \texttt{ROS}
for onboard communication and control.
The centralized planning and control algorithm runs on a laptop (Intel Core i7-1280P),
which wirelessly transmits the velocity commands $\mathbf{u}_n$ in~\eqref{eq:control} to the robots,
and executed by their onboard velocity controllers.
Objects are made of cardboard, each weighing approximately~$1\,\mathrm{kg}$,
with a ground friction coefficient of~$0.5$ and a side friction coefficient of~$0.2$.
  These parameters were \emph{not} meticulously calibrated, e.g., using force sensors,
  but are sufficient for the algorithm to complete the target tasks,
  as the kinematic-level control is relatively insensitive to those force-related parameters.
With above setup, two scenarios were tested:
(I) a T-shaped and an L-shaped object placed on opposite sides of
a narrow passage formed by bar-shaped obstacles,
with four robots swapping their positions under strict task ordering constraints;
(II) an L-shaped object, a long rectangular object,
and a T-shaped object arranged in a triangular layout,
where four robots rotate the objects counter-clockwise around a center obstacle.

\subsubsection{Results}\label{subsec:exp-results}

As shown in Fig.~\ref{fig:hardware},
the proposed method successfully completes the pushing tasks in both scenarios.
In Scenario-I, MAPF generates two collision-free trajectories in~$8.2$s,
decomposed into~$5$ subtasks in~$0.5$s.
Robots are divided into two subgroups in~$0.2$s for task assignment:
one subgroup pushes the T-shaped object upwards,
while the other pushes the L-shaped object right.
By~$t=40$s, the T-shaped object reaches its target,
and the L-shaped object begins its second subtask.
All objects reach their goals by~$t=85$s with~$7$ modes.
In Scenario-II, MAPF generates three timed trajectories in~$1.2$s,
decomposed into~$5$ subtasks in~$0.6$s.
Robots are divided into two subgroups in~$0.2$s.
In particular, the L-shaped object subtask~$(2,1)$ is assigned to one subgroup,
and the rectangular object subtask~$(3,1)$ is assigned to the other.
By~$t=45$s, the rectangular object has reached its target position,
clearing space for the T-shaped object.
The L-shaped and rectangular objects are pushed to their targets by~$t=45$s,
and the remaining tasks are completed by~$t=65$s with~$6$ modes.
The real-time progress and velocity control inputs are recorded in Fig.~\ref{fig:hardware}.
Note that the actual execution time fluctuates
due to the poor motion of Mecanum wheels in certain directions around~$45^\circ$,
causing drifting and slipping.
However, the partial ordering of subtasks and the online adaptation scheme
ensure the system consistently completes the tasks.
More details can be found in the attached videos.

%% file: contents/conclusion.tex
\section{Conclusion} \label{sec:conclusion}
This work proposes a combinatorial-hybrid optimization scheme for
synthesizing collaborative pushing strategies
for multiple robots and multiple objects.
The scheme consists of two interleaved layers:
the decomposition and assignment of pushing subtasks;
and the hybrid optimization of pushing modes and forces
for each subtask.
Moreover, a diffusion-based predictor is trained to
accelerate the generation of pushing modes during hybrid optimization.
It is shown to be scalable and effective for general-shaped objects
in cluttered scenes.
The completeness and feasibility of our algorithm have been demonstrated
both theoretically and experimentally.
Future work includes
partial knowledge of the objects, distributed coordination schemes,
and other collaborative tasks with
dynamical and physical constraints among the robots.

%% file: contents/ack.tex
\section*{Acknowledgment}
The authors would like to thank Zhengyu Yang for his help on the hardware experiments.

%% file: contents/appendix.tex
\appendix

\begin{table}[ht]\
\color{black}
\footnotesize 
\centering
\caption{Nomenclature Table}
\label{tab:terminology}
\vspace{-2mm}
\begin{tabularx}{\linewidth}{p{2.2cm}Xl}
\toprule
\textbf{Term} & \textbf{Definition} & \textbf{Reference} \\
\midrule
Pushing Mode ($\boldsymbol{\xi}$) & Contact points and forces between robots and the target object. & Sec.~\ref{subsec:ws} \\
\addlinespace
MAPF Path ($\mathfrak{S}_m$) & Collision-free timed trajectory. & Eq.~\eqref{eq:m-path} \\
Subtask ($\mathfrak{S}^k_m$) & $k$-th segment of object path. & Eq.~\eqref{eq:m-seg} \\
Partial Order ($\preceq$) & Temporal ordering between the subtasks. & Def.~\ref{def:ordering} \\
Task Plan ($\boldsymbol{\tau}_i$) & Timed sequence of subtasks as the local plan for robot~$i$. & Def.~\ref{def:plan} \\
Subgroup ($\mathcal{N}^k_m$) & Robot coalition for subtask. & Sec.~\ref{subsubsec:assign} \\
\addlinespace
Keyframe ($\kappa_\ell$) & Critical system state. & Sec.~\ref{subsubsec:hybrid-search} \\
Hybrid Plan ($\vartheta$) & A sequence of keyframes and pushing modes. & Eq.~\eqref{eq:iter-sample} \\
Arc Segment ($\varrho_\ell$) & Trajectory between keyframes. & Sec.~\ref{subsubsec:hybrid-search} \\

\addlinespace
Primitive Plan ($\widehat{\vartheta}$) & Feasible hybrid plan as sequence of keyframes and modes. & Alg.~\ref{alg:hybrid} \\
Diffusion Model & Diffusion-based neural network for generating hybrid plans. & Sec.~\ref{subsubsec:diffusion} \\
Primitive Lib. ($\mathcal{X}$) & Library that contains verified hybrid plans for different subtasks. & Alg.~\ref{alg:hybrid} \\
Mode Lib. & Library of verified modes. & Alg.~\ref{alg:hybrid} \\
\bottomrule
\end{tabularx}
\vspace{-4mm}
\end{table}


\subsection{{Modeling and Mode Feasibility}}\label{app:model}
  \subsubsection{Pushing Modes and Coupled Dynamics}\label{app:physics}
  Given an object $m$ and a set of robots $\mathcal{N}_m$,
  working at a particular pushing mode $\xi$,
  the robots can apply pushing forces in different directions
  with different magnitude.
  Denote by~$\mathbf{f}_1 \mathbf{f}_2 \cdots \mathbf{f}_{N_m}$,
  where~$\mathbf{f}_n\in \mathbb{R}^2$ is the contact force of robot~$R_n$ at
  contact point~$\mathbf{c}_n$, $\forall n\in \mathcal{N}_m$.
  Furthermore, each force~$\mathbf{f}_n$ can be decomposed
  in the directions of the normal vector~$\mathbf{n}_n$ and the tangent vector~$\bm{\tau}_{n}$
  w.r.t.~the object surface at the contact point~$\mathbf{c}_n$,
  i.e.,~$\mathbf{f}_n\triangleq \mathbf{f}^{\texttt{n}}_n + \mathbf{f}^{\texttt{t}}_n
  \triangleq f^{\texttt{n}}_n \mathbf{n}_n + f^{\texttt{t}}_n \mathbf{n}^{\perp}_n$.
  Due to the Coulomb law of friction see~\cite{kao2016contact},
  it holds that:
  \begin{equation}\label{eq:force-limit}
  0\leq f^{\texttt{n}}_n \leq f_{n,\max};
  \; 0\leq |f^{\texttt{t}}_n| \leq \mu^{\texttt{c}}_{m} f^{\texttt{n}}_n,
  \end{equation}
  where~$f_{n,\max}> 0$ is the maximum force each robot can apply;
  and~$\mu^{\texttt{c}}_{m}$ is the coefficient of lateral friction defined earlier.
  Then, these decomposed forces can be re-arranged by:
  \begin{equation}\label{eq:F-xi}
  \mathbf{F}_{\xi}\triangleq (\mathbf{F}^{\texttt{n}}_{\xi},\,
  \mathbf{F}^{\texttt{t}}_{\xi})
  \triangleq (f^\texttt{n}_{1},\cdots,f^\texttt{n}_{\sss{N_m}},
  f^\texttt{t}_{1},\cdots,f^\texttt{t}_{\sss{N_m}})\in\mathbb{R}^{\sss{2N_m}},
  \end{equation}
  and further~$\mathcal{F}_{\xi}\triangleq \{\mathbf{F}_{\xi}\}$
  denotes the set of all forces within each mode~$\boldsymbol{\xi}\in \Xi_m$.
  Furthermore, the combined generalized
  force~$\mathbf{q}_{\xi}\triangleq (\mathbf{f}^\star,\chi^\star)$
  as also used in~\cite{lynch1992manipulation} is given by:
  \begin{equation}\label{eq:generalized_force}
      \mathbf{f}^\star\triangleq \textstyle\sum_{n=1}^{N_m}{\mathbf{f}_n};\;
      \chi^\star\triangleq \sum_{n=1}^{N_m}{(\mathbf{c}_n-\mathbf{x}_n)\times \mathbf{f}_n},
  \end{equation}
  where~$\times$ is the cross product and~$\chi^\star$ is the resulting torque from all robots.
  It can be written in matrix form~$\mathbf{q}_{\xi}\triangleq \mathbf{J}\mathbf{F}_{\xi}$,
  where $\mathbf{J}\triangleq \nabla_{\sss{\mathbf{F}_{\xi}}} \mathbf{q}_{\xi}$ is
  a~$3\times 2N_m$ Jacobian matrix.
  Similarly, let $\mathrm{Q}_{\xi}\triangleq \{\mathbf{q}_{\xi}\}$
  denote the set of all \emph{allowed} combined generalized forces
  within each mode~$\boldsymbol{\xi} \in \Xi_m$.
  The coupled dynamics of the object and robots
  can be described as follows:
  \begin{subequations}\label{eq:multi_body}
    \begin{align}
      \mathbf{M}_m\dot{\mathbf{p}}_m
      &= \mathbf{q}_{\mu_m}+\mathbf{q}_{\xi_m}
      = Q^m_{\mu}(\mathbf{p}_m) +
      \textstyle\sum_{n\in\mathcal{N}_m} \mathbf{q}_{m,n};
      \label{eq:object_equation}\\
      \mathbf{M}_n \dot{\mathbf{p}}_n
      &= Q^n_{\texttt{drv}}(\mathbf{u}_n,\mathbf{s}_n,\mathbf{p}_n) - \mathbf{q}_{m,n},
      \label{eq:robot_equation}
    \end{align}
  \end{subequations}
  where~$\mathbf{M}_m \triangleq \texttt{diag}(\mathsf{M}_m,\mathsf{M}_m,\mathsf{I}_m)$;
  $\mathbf{q}_{m,n}$ is the generalized pushing force in~\eqref{eq:generalized_force}
  applied by robot~$R_n$ on object~$\Omega_m$;
  and~$\mathbf{q}_{m,\mu}\triangleq Q^m_{\mu}(\mathbf{p}_m)=(\mathbf{f}_m,\chi_m)$
  is the ground friction, determined by the velocity~$\mathbf{p}_m$
  and object intrinsics.
  \eqref{eq:object_equation} models the object's motion
  under external forces \(\mathbf{q}_{\xi}\) and \(\mathbf{q}_\mu\) within mode \(\xi\),
  while \eqref{eq:robot_equation} describes the robot's motion under the control inputs \(\mathbf{u}_n\)
  and the reaction force from the object.
  These equations are instrumental for system modeling and subsequent analysis,
  while the simulation is handled by the physics engine.
  Denote by~$\boldsymbol{\xi}_m(t)\triangleq (\xi_m(t), \mathbf{q}_{\xi_m}(t), \mathcal{N}_m(t))$
  the contact points, pushing forces and participants for object~$m\in\mathcal{M}$ at time~$t\geq 0$.
  
  \begin{figure}
    \centering
    \includegraphics[width=0.95\linewidth]{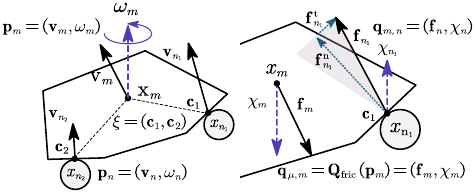}
    \vspace{-2mm}
    \caption{{Illustration of a pushing mode
        and relevant notations in Sec.~\ref{app:physics}.}}
    \label{fig:contact}
    \vspace{-4mm}
  \end{figure}

  \subsubsection{Quasi-static Analyses}\label{app:quasi-static}
  The friction force~$Q_\mu(\mathbf{p}_m)$ in~\eqref{eq:object_equation}
  lacks a closed-form expression.
  As also adopted in~\cite{lynch1992manipulation,tang2024collaborative},
  the quasi-static analyses assume that the motion of the target is sufficiently slow,
  such that its acceleration is approximately zero and the inertia forces can be neglected.
  Consequently, given the desired velocity~$\mathbf{p}_m^\star$ of the object,
  the generalized friction force~$\mathbf{q}_{m,\mu}$ is computed as:
  \begin{equation}\label{eq:friction_force}
    \mathbf{q}_{\mu_m} \triangleq \widetilde{Q}_{\mu}^m(\mathbf{p}_m^\star)=
    -\|\mathbf{D}_{1}\mathbf{D}_{2}\mathbf{p}_m\|^{-1}_{_2}\mathbf{D}_{2}\mathbf{p}_m^\star,
  \end{equation}
  where $\widetilde{Q}_\mu^m(\cdot)$ approximates $Q_\mu(\cdot)$ in~\eqref{eq:object_equation};
  $\mathbf{D}_{1}\triangleq
  \text{diag}(f_{\sss{\max}},f_{\sss{\max}},m_{\sss{\max}})^{-1}$,
  $\mathbf{D}_{2}\triangleq \text{diag}(1,1,m_{\max}^2/f_{\max}^2)$;
  $f_{\max}$ and~$m_{\max}$ are the maximum ground friction and moment
  of the target object~$\Omega_m$.
  Furthermore,
  assuming no slipping occurs during the pushing process,
  and the robot continuously applies a net force that exactly counteracts the frictional force, i.e.,~$\mathbf{q}_{m,\xi} = -\mathbf{q}_{\mu_m}$,
  the object can be pushed at a constant body-frame velocity~$\mathbf{p}_m^{\bsss}$.
  Under this condition, the resulting trajectory forms a circular arc, expressed as:
  $\Delta \mathbf{s}(t)\triangleq \Big{(}\int_{t=0}^{\bar{t}} \mathbf{Rot}(\omega t+ \psi_0) dt\Big{)}\,
  \mathbf{p}_m^{\bsss}$,
  where~$\mathbf{Rot}(\cdot)$ denotes the rotation matrix,
  $\omega$ is the angular velocity in $\mathbf{p}^{\bsss}$,
  and $\psi_0$ is the initial orientation of the object.
  Conversely,
  given any two states,
  $\mathbf{s}_{\ell}$ and $\mathbf{s}_{\ell+1}$,
  the corresponding arc trajectory $\varrho_{\ell}=\overgroup{\mathbf{s}_\ell \mathbf{s}_{\ell+1}}$
  can be uniquely determined,
  along with the body-frame velocity~$\mathbf{p}^{\bsss}_{\varrho}$ that generates it.

  \subsubsection{{Feasibility of Pushing Modes}}\label{app:feasibility}
  We define the \textit{primary feasibility loss}~$\mathrm{J}^m_{\texttt{F}}$ to
  evaluate whether the pushing forces~$\mathbf{q}_{m,\xi}$
  are sufficient to counteract frictional forces
  while maintaining the object's motion
  at the desired body-frame velocity~$\mathbf{p}^{\bsss}$:
  \begin{equation}\label{eq:feasibility-loss}
    \mathrm{J}^m_{\texttt{F}}(\boldsymbol{\xi},\,\mathbf{p}^{\bsss}) \triangleq
    \underset{\mathbf{q}_{m,\xi} \in \mathrm{Q}_{m,\xi}}{\textbf{min}}
    \left\| \mathbf{q}_{m,\xi} + Q_{\mu}^m(\mathbf{p}^{\bsss}) \right\|_{1},
  \end{equation}
  where~$\|\|_1$ is the first norm.
  To account for the robustness of the mode against perturbations in velocity direction,
  we introduce the \textit{multi-directional feasibility loss}~$\mathrm{J}^m_{\texttt{MF}}$,
  which aggregates the primary loss over a set of basis velocities~$\mathcal{D}$:
  \begin{equation}\label{eq:multi-directional-feasibility}
    \mathrm{J}^m_{\texttt{MF}}(\boldsymbol{\xi},\,\mathbf{p}^{\bsss}) \triangleq
    \sum_{\mathbf{p}_d \in \mathcal{D}} w_d \cdot \mathrm{J}^m_{\texttt{F}}(\boldsymbol{\xi},\,\mathbf{p}_d),
  \end{equation}
  where~$\mathcal{D}$ denotes a set of basis velocities that span the generalized velocity space,
  including the desired direction~$\mathbf{p}^{\bsss}$.
  Each basis direction~$\mathbf{p}_d$
  is associated with a weight~$w_d$, with the primary direction assigned the highest weight.
  This loss is introduced as a soft measure to account for infeasibility and uncertainty in certain scenarios.
  A lower~$\mathrm{J}^m_{\texttt{MF}}$ indicates greater robustness of the pushing mode~$\boldsymbol{\xi}$, 
  i.e., it can tolerate perturbations in velocity direction while maintaining feasibility.
  {
  \begin{definition}
    (I) A mode~$\boldsymbol{\xi}$ is \textit{force-feasible} for a target velocity $\mathbf{p}^{\bsss}$ \textit{if and only if}
    $\mathrm{J}^m_{\texttt{F}}(\boldsymbol{\xi},\,\mathbf{p}^{\bsss}) = 0$;
    (II) A mode~$\boldsymbol{\xi}$ is \textit{practically feasible} for a target velocity $\mathbf{p}^{\bsss}$ \textit{if and only if}
    the robot can stably push the object at velocity $\mathbf{p}^{\bsss}$. \hfill $\blacksquare$
  \end{definition}
  }
  Force feasibility is necessary but not sufficient for practical feasibility.
  For example, robots without force sensing cannot precisely apply desired forces
  via position control alone.
  Thus, given a pushing mode~$\boldsymbol{\xi}$ and a target velocity~$\mathbf{p}^{\bsss}$,
  the feasibility check is performed in two stages:
  {(I) Force feasibility}:
  Verify if $\mathrm{J}^m_{\texttt{F}}(\boldsymbol{\xi},\,\mathbf{p}^{\bsss}) = 0$ holds;
  {(II) Practical feasibility}:
  If force-feasible, verify the mode in simulation
  by tracking the arc trajectory $\varrho_\mathbf{p}$ generated by $\mathbf{p}$.
  If successful, the mode is considered practically feasible.
  Additionally, for real-world execution,
  a mode library is maintained to record the mode and observed tracking error.
  This helps bridge the gap between
  simulation-based feasibility checks and physical execution.

\subsubsection{Mode Sufficiency}\label{app:sufficient}

  The subgroup of robots~$\mathcal{N}_m$ is said to be \emph{mode-sufficient} for pushing object~$\Omega_m$
  if there exists a set of velocities~$\mathrm{P}^\star_m\triangleq \{\mathbf{p}^\star_j\}$ such that:
{  (I) $\mathrm{P}^\star_m$ can positively span the $\mathbb{R}^3$ space;
  and (II) the robots in~$\mathcal{N}_m$ can push object~$\Omega_m$
  in each velocity~$\mathbf{p}_j^\star\in \mathrm{P}^\star_m$,
  i.e., there exists a practically feasible mode~$\boldsymbol{\xi}_j$ for each velocity~$\mathbf{p}_j^\star$.}
The set~$\mathrm{P}^\star_m$ satisfying the above sufficient conditions
  is referred to as the set of {feasible velocities} for the system.
  Verification of this condition can be done by iteratively expand the set of feasible velocities,
  i.e., by selecting new velocity direction~$\mathrm{p}$ and generate modes by~\eqref{eq:mode_gen}.
  If a feasible mode is found, this velocity is added to the set~$\mathrm{P}^\star_m$.
  This process is repeated until when the set~$\mathrm{P}^\star_m$ can span the~$\mathbb{R}^3$ space
  or the set of velocities is exhausted.

\subsection{{Proof of Lemmas and Theorems}}\label{app:proof}
\begin{proof}
  \textbf{of Lemma~\ref{lemma:decompose}}.
    Assume that after certain iterations,
    the largest splitting instance~$t^\star_m$ remains unchanged for all objects,
    and there exists~$m\in \mathcal{M}$ such that~$t_{m}^\star<t_L$.
    Let $m^\star \triangleq \textbf{argmin}_{m\in \mathcal{M}}\{t_m^\star\}$.
    (I) If $\forall m'$, $t^\star_m< t^\star_{m'}$,
    the next splitting instance~$t^{\texttt{s}}_{m^\star}$ is determined by~\eqref{eq:split},
    which implies that $t^{\texttt{s}}_{m^\star} \geq t^\star_{m'} > t^\star_{m^\star}$.
    Thus $t^{\texttt{c}}_{m'}$ will increase, contradicting the assumption.
    {
    (II) If there exists~$m'$ such that~$t^\star_{m'}=t^\star_{m^\star}$,
    which implies that two objects collide at same time~$t^\star_{m^\star}$,
    which contradicts collision-free assumption of the timed paths.
    }
    Thus, $t^\star_m$ can reach $t_L$ in finite steps and the algorithm terminates.
    Moreover, assume that there exists a loop in the ordering
    i.e., $\mathfrak{S}^{k}_{m}\preceq\cdots\preceq\mathfrak{S}^{k}_{m}$.
    If segment $\mathfrak{S}^{k}_{m}\preceq\mathfrak{S}^{k'}_{m'}$,
    then it holds that~{$t^{k,\texttt{c}}_{m}\leq t^{k',\texttt{c}}_{m'}$}.
    Following~\eqref{eq:split},
    $\mathfrak{S}^{k'}_{m'}$ can only be created
    after the segment $\mathfrak{S}^{k}_{m}$ has been created
    and removed from $\widetilde{\mathfrak{S}}_m$.
    Thus, $\mathfrak{S}^{k}_{m}$ must be created before itself,
    which contradicts the assumption,
    the partial ordering is strict without loops.
  \end{proof}

  \begin{proof}
    \textbf{of Lemma~\ref{lemma:ordered_execution}}.
    To begin with, the segments of the same object are
    followed according to the sequence in~$\overline{\mathfrak{S}}_m$,
    i.e., sequentially from~$\mathfrak{S}^{1}_m$ to~$\mathfrak{S}^{K_m}_m$.
    {Then, given any two segments
    $\mathfrak{S}^{k_1}_{m_1}$ and~$\mathfrak{S}^{k_2}_{m_2}$
    of different objects $m_1$ and $m_2$,
    consider the following two cases:
    if two segments are not partially ordered,
    objects $m_1$ and $m_2$ can be moved concurrently;
    otherwise,
    if $\mathfrak{S}^{k_1}_{m_1} \preceq \mathfrak{S}^{k_2}_{m_2}$,
    the second condition requires that the subsequent object~$m_2$
    can only be moved after the preceding segment of object~$m_1$ has been traversed.}
    In this way, if each segment is traversed within a bounded time,
    each object can reach its goal state without collision.
  \end{proof}

  \begin{proof}
  {
  (Sketch) \textbf{of Lemma~\ref{lemma:hybrid}}.
  The key insight is that
  any arc in collision will be split into two shorter arcs,
  until all the arcs are collision-free.
  For these collision-free arcs,
  the iterative sampling procedure
  will try to find intermediate keyframes and feasible modes.
  If the iterative sampling fails,
  the collision-free arc will be further split.
  When the arc $\varrho_{\ell}$ is sufficiently short,
  i.e. $|\varrho_{\ell}|<\epsilon$,
  it can be approximated by a sequence of three arcs.
  Specifically, the arc $\varrho_{\ell}$ can be generated by
  velocity $\mathbf{p}_{\ell}$ with time duration $t_{\ell}$.
  With the mode sufficiency assumption,
  any velocity $\mathbf{p}_{\ell}$ can be positively decomposed
  into three primitive velocities $\mathbf{p}^\star_j$,
  where~$j\in\{1,2,3\}$, i.e.,
  $\mathbf{p}_{\ell} = \sum_{j=1}^{3} \lambda_j^\star \mathbf{p}^\star_j$,
  for $\lambda_j^\star \geq 0$ and $j\in\{1,2,3\}$.
  Consider the sequence of arc motion
  $(\mathbf{p}^\star_1,t_1)(\mathbf{p}^\star_2,t_2)(\mathbf{p}^\star_3,t_3)$
  that starts from~$\mathbf{s}_{\ell}$,
  where $t_j=\lambda_j^\star t$ is the duration of velocity $\mathbf{p}^\star_j$.
  The Hausdorff distance between the trajectory sequential arc motion and the original arc $\varrho_{\ell}$
  can be proved to be bounded by $\mathcal{O}(\epsilon)$,
  as detailed in the supplementary material.
  As each primitive velocity has a feasible mode,
  this sequence of arc motion can be replaced by the original arc $\varrho_{\ell}$
  to form a feasible hybrid plan.
  Consequently, the search depth is bounded by~$|\overline{\mathfrak{S}}_m^k|/\epsilon$
  with~$|\overline{\mathfrak{S}}_m^k|$ being the total length of the path segment,
  yielding that the termination condition is reached in finite steps.
  Thus, the algorithm is guaranteed to find a feasible solution.
  }
  \end{proof}
  
  \begin{proof}
    {
  \textbf{of Theorem~\ref{theo:all}}.
  Under condition C2 in Table~\ref{tab:assumption},
  the collision-free timed paths $\{\mathfrak{S}_m,\forall m\in \mathcal{M}\}$
  can be found by MAPF algorithms.
  Alg.~\ref{alg:segments}
  decomposes the paths $\{\mathfrak{S}_m,\forall m\in \mathcal{M}\}$
  into $\{\overline{\mathfrak{S}}_m\}$ that satisfies strict partial ordering,
  with a guarantee on convergence in Lemma~\ref{lemma:decompose}.
  Then, the segments are assigned to the robots with a horizon of $H$,
  while ensuring that the partial ordering are respected,
  and furthermore those segments can be traversed without collisions by Lemma~\ref{lemma:ordered_execution}.
  Meanwhile, under the condition C3,
  each robot subgroup assigned to a subtask must satisfy
  the mode-sufficient condition,
  which guarantees that
  the hybrid strategy for each object is feasible by Lemma~\ref{lemma:hybrid}.
  Lastly,
  under the proposed motion controller,
  each robot can track any given arc motion of the object,
  and apply the required bounded forces.
  Thus, all subtasks can be accomplished, which in turns
  ensures that the overall pushing task for
  each object can be fulfilled.}
  \end{proof}